\documentclass[10pt]{article} % For LaTeX2e
\usepackage[accepted]{tmlr}
\usepackage{amsmath,amsfonts,bm}

\def\eqref#1{equation~\ref{#1}}
\def\1{\bm{1}}

\DeclareMathAlphabet{\mathsfit}{\encodingdefault}{\sfdefault}{m}{sl}
\SetMathAlphabet{\mathsfit}{bold}{\encodingdefault}{\sfdefault}{bx}{n}

\DeclareMathOperator{\Tr}{Tr}

\usepackage{hyperref}
\usepackage{url}
\usepackage{graphicx}
\usepackage{subcaption}
\usepackage{amsthm}
\usepackage{multirow}
\usepackage{wrapfig}
\usepackage{amssymb} % For checkmarks and crosses
\usepackage{colortbl} % For row and column colors
\usepackage{pifont} % For symbols like dingbats
\usepackage[linesnumbered,ruled,vlined]{algorithm2e}\usepackage{caption}
\usepackage{wrapfig}
\usepackage{tabularx,ragged2e}
\newtheorem{lemma}{Lemma}
\usepackage{ulem}
\newtheorem{remark}{Remark}
\newtheorem{definition}{Definition}[section]

\newcommand{\bcheck}{{\color[HTML]{007A33}\ding{51}}} % Bold green checkmark
\newcommand{\bcross}{{\color[HTML]{CC0000}\ding{55}}} % Bold red cross

\newcommand{\ourmethod}{Sep-SpectralNet}

\title{Generalizable Spectral Embedding with an Application to UMAP}

\author{\name Nir Ben-Ari \email nirnirba@gmail.com \\
      \addr Department of Computer Science\\
      Bar-Ilan University \\
      \AND
      \name Amitai Yacobi \email amitaiyacobi@gmail.com \\
      \addr Department of Computer Science\\
      Bar-Ilan University
      \AND
      \name Uri Shaham \email uri.shaham@biu.ac.il\\
      \addr Department of Computer Science\\
      Bar-Ilan University}

\def\month{07}  % Insert correct month for camera-ready version
\def\year{2025} % Insert correct year for camera-ready version
\def\openreview{\url{https://openreview.net/forum?id=8cuQwztCKk}} % Insert correct link to OpenReview for camera-ready version

\begin{document}

\maketitle

\begin{abstract}
Spectral Embedding (SE) is a popular method for dimensionality reduction, applicable across diverse domains. Nevertheless, its current implementations face three prominent drawbacks which curtail its broader applicability: generalizability (i.e., out-of-sample extension), scalability, and eigenvectors separation. Existing SE implementations often address two of these drawbacks; however, they fall short in addressing the remaining one.
In this paper, we introduce \textit{\ourmethod} (eigenvector-separated SpectralNet), a SE implementation designed to address \textit{all} three limitations.
\ourmethod\ extends SpectralNet with an efficient post-processing step to achieve eigenvectors separation, while ensuring both generalizability and scalability. This method expands the applicability of SE to a wider range of tasks and can enhance its performance in existing applications.
We empirically demonstrate \ourmethod's ability to consistently approximate and generalize SE, while maintaining SpectralNet's scalability. Additionally, we show how \ourmethod\ can be leveraged to enable generalizable UMAP visualization.\footnote{\ourmethod: \url{https://github.com/shaham-lab/GrEASE}; NUMAP: \url{https://github.com/shaham-lab/NUMAP}}
\end{abstract}

\section{Introduction}
\label{sec:intro}
Spectral Embedding (SE) is a popular non-linear dimensionality reduction method \citep{laplacianEigenmaps, diffusionMaps}, finding extensive utilization across diverse domains in recent literature. Notable applications include UMAP \citep{umap} (the current state-of-the-art visualization method), Graph Neural Networks \citep{zhang2021eigen, beaini2021directional, defferrard2016convolutional}, signal propagation on graphs \citep{park2022multiscale, klaine2017survey} and analysis of proteins \citep{campbell2015laplacian, zhu2021fiedler}.
The core of SE involves a projection of the samples into the space spanned by the leading eigenvectors of the Laplacian matrix (i.e., those corresponding to the smallest eigenvalues), derived from the pairwise similarities between the samples.
It is an expressive method which is able to preserve the global structure of high-dimensional input data, underpinned by robust mathematical foundations \citep{laplacianEigenmaps, katz2019alternating, lederman2018learning, ortega2018graph}.

Despite the popularity and significance of SE, current implementations suffer from three main drawbacks: (1) \textit{Generalizability} - the ability to directly embed a new set of test points after completing the computation on a training set (i.e., out-of-sample extension); (2) \textit{Scalability} - the ability to handle a large number of samples within a reasonable time-frame; (3) \textit{Eigenvectors separation} - the ability to output the \textit{basis} of the leading eigenvectors \((v_2, \dots, v_{k+1})\), rather than only the space spanned by them.
These three properties are crucial for modern applications of SE in machine learning. While most SE implementations address two of these three limitations, they often fall short in addressing the remaining one (see Tab. \ref{tab:methods} and Sec. \ref{sec:related_work}).

Notably, eigenvector separation has attracted considerable attention in recent years \citep{pfau2018spectral, gemp2020eigengame, deng2022neuralef, lim2022sign}. This work focuses on this property, which is a key aspect underlying applications such as Fiedler vector, Diffusion maps, and Diffusion analysis. The latter two involve examining the evolution of random walks on graphs across different time scales. This framework enables a wide range of applications: analyzing molecular transitions \citep{glielmo2021unsupervised, chiavazzo2017intrinsic}, sampling rare protein transitions \citep{ghamari2024sampling}, uncovering latent structures in disordered materials \citep{hardin2024revealing}, studying the global organization of cortical features in various neuropsychiatric conditions \citep{park2021differences, park2022multiscale, dong2020compression}, supporting enhanced-sampling techniques \citep{zheng2011delineation, zheng2013rapid}, and facilitating dimensionality reduction in self-organizing networks \citep{klaine2017survey}. Eigenvector separation is required for those applications to enable scaling each diffusion component.

\cite{spectralnet} presented SpectralNet, which tackles the scalability and generalizability limitations of Spectral Clustering (SC), a key application of SE. However, we prove that due to a rotation and reflection ambiguity in its loss function, SpectralNet cannot directly be adapted for SE in general, as it cannot separate the eigenvectors. In this paper, we present an eigendecomposition based post-processing procedure to resolve the eigenvectors separation issue in SpectralNet, thereby, extending SpectralNet into a scalable and generalizable implementation of SE, which we call \textit{\ourmethod} (eigenvector-separated SpectralNet). \ourmethod's ability to separate the eigenvectors, while maintaining the generalizability and scalability of SpectralNet, offers a pathway to enhance numerous existing applications of SE, that require eigenvector separation, and provides a foundation for developing new applications.

One such application is UMAP \citep{umap}, a widely used visualization method whose ability to preserve global structure relies heavily on its Spectral Embedding (SE) initialization \citep{kobak2021initialization}. Although Parametric UMAP (P. UMAP) was proposed to address UMAP’s lack of generalizability \citep{sainburg2021parametric}, it omits SE, limiting its ability to replicate UMAP’s structural fidelity. Despite this limitation, P. UMAP has gained traction in various domains \citep{xu2023robust, eckelt2023visual, leon2021data, xie2023integrative, yoo2022online}, underscoring the need for a generalizable alternative that retains UMAP’s structural strengths. To this end, we introduce an extension of \ourmethod\ that integrates SE-based initialization with the UMAP loss, named NUMAP, preserving global structure while enabling generalization. We show that eigenvector separation is required for NUMAP's performance. As shown empirically (Sec. \ref{sec:exps}), this extension enhances UMAP's applicability to dynamic settings such as online learning and time-series data visualization.

Our key contributions are:
(1) We introduce \ourmethod, a novel approach for approximate Spectral Embedding (SE) that jointly addresses scalability, generalizability, and eigenvector separation.
(2) We present an application of \ourmethod to generalizable UMAP. (3) We propose a new evaluation method to measure global structure preservation.

\section{Related Work}
\label{sec:related_work}

\begin{table}[t]
\centering
\renewcommand{\arraystretch}{1.5}
\begin{tabular}{|l|c|c|c|}
\hline
\rowcolor[HTML]{EFEFEF} 
\textbf{Method}       & \textbf{Generalizability} & \textbf{Scalability} & \textbf{Eigenvector Separation} \\ \hline
LOBPCG              & \bcross & \bcheck & \bcheck \\ \hline
SpectralNet              & \bcheck & \bcheck & \bcross \\ \hline
DiffusionNet              & \bcheck & \bcross & \bcheck \\ \hline
\textbf{\ourmethod\ (ours)}              & \bcheck & \bcheck & \bcheck \\ \hline
\end{tabular}
\caption{\textbf{\ourmethod\ is the only method to have the three desired properties of SE implementation.} Comparison between key SE methods via their ability to generalize to unseen samples, scale to large datasets and separate the eigenvectors.}
\label{tab:methods}
\end{table}

Current SE implementations typically address two out of its three primary limitations: generalizability, scalability, and eigenvector separation (Tab. \ref{tab:methods}). Below, we outline key implementations that tackle each pair of these challenges. Following this, we discuss recent works related to eigenvectors separation and generalizable visualizations techniques.

\paragraph{Scalable with eigenvectors separation.} Popular implementations of SE are mostly based on sparse matrix decomposition techniques (e.g., ARPACK \citep{lehoucq1998arpack}, AMG \citep{brandt1984algebraicAMG}, LOBPCG \citep{benner2011locallyLOBPCG}). These methods are relatively scalable, as they are almost linear in the number of samples. Nevertheless, their out-of-sample extension is far from trivial. Usually, it is done by out-of-sample extension (OOSE) methods such as Nystr\"{o}m \citep{nystrom1930praktische} or Geometric Harmonics \citep{coifman2006geometric, lafon2006data}. However, these methods provide only local extension (i.e., near existing training points), and are both computationally and memory restrictive, as they rely on computing the distances between every new test point and all training points.
% GrEASE not only generalizes to unseen points with a single feed-forward operation but is also empirically faster than existing methods.

\paragraph{Scalable and generalizable.} Several approaches to spectral clustering (SC) approximate the space spanned by the first eigenvectors of the Laplacian matrix, which is sufficient for clustering purposes, and can also benefit other specific applications. For example, SpectralNet \citep{spectralnet} leverages deep neural networks to approximate the first eigenfunctions of the Laplace-Beltrami operator in a scalable manner, thus also enabling fast inference of new unseen samples. BASiS \citep{streicher2023basis} achieves these goals using affine registration techniques to align batches. However, these methods' inability to separate the eigenvectors prevents their use in many modern applications.

\paragraph{Generalizable with eigenvectors separation.} Another proposed approach to SE is DiffusionNet \citep{mishne2019diffusion}, a deep-learning framework for generalizable Diffusion Maps embedding \citep{diffusionMaps}, which is similar to SE. However, the training procedure of the network is computationally expensive, therefore restricting its usage for large datasets.

In contrast, we introduce \ourmethod, which addresses all three limitations - generalizes the separated eigenvectors to unseen points with a single feed-forward operation, while maintaining SpectralNet's scalability.

\paragraph{Eigenvectors separation.} Extensive research has been conducted on the eigenvectors separation problem, both within and beyond the spectral domain \citep{lim2022sign, ma2024laplacian}. Some rotation criteria such as ICA and VARIMAX are well known, but regarding the spectral domain, they do not yield the natural separation, i.e., the true eigenvectors. Recent spectral approaches remain constrained computationally, both by extensive run-time and memory consumption. For example, \cite{pfau2018spectral} proposed a solution to this issue by masking the gradient information from the loss function. However, this approach necessitates the computation of full Jacobians at each time step, which is highly computationally intensive.
\cite{gemp2020eigengame} employs an iterative method to learn each eigenvector sequentially. Namely, they learn an eigenvector while keeping the others frozen. This process has to be repeated \(k\) times (where \(k\) is the embedding dimension), which makes this approach also computationally expensive. \cite{deng2022neuralef} proposed an improvement of the latter, by parallel training of \(k\) NNs. However, as discussed in their paper, this approach becomes costly for large values of \(k\). Furthermore, it necessitates retaining \(k\) trained networks in memory, which leads to significant memory consumption. \cite{chen2022specnet2} proposed a post-processing solution to this problem using the Rayleigh-Ritz method. However, this approach involves the storage and multiplication of very large dense matrices, rendering it impractical for large datasets. In contrast, \ourmethod\ offers an efficient one-shot post-processing solution to the eigenvectors separation problem.

\paragraph{Generalizable visualization.} Several works have attempted to develop parametric approximations for non-parametric visualization methods, in addition to Parametric UMAP (P. UMAP) \citep{sainburg2021parametric}. Notable examples include \citep{van2009learning}, \citep{kawase2022parametric} and \citep{damrich2022t}, which use NNs to make t-SNE generalizable, and \citep{schofield2021using}, which aims to make UMAP more interpretable. However, P. UMAP has demonstrated superior performance. NUMAP presents a method to surpass P. UMAP in terms of global structure preservation.

\section{Preliminaries} \label{sec:pre}
In this section, we begin by providing the fundamental definitions that will be used throughout this work. Additionally, we briefly outline the key components of UMAP and P. UMAP.

\subsection{Spectral Embedding}
\label{sec:pre_se}
Let \(\mathcal{X} = \{x_1, \dots, x_n\} \subseteq \mathbb{R}^d\) denote a collection of unlabeled data points drawn from some unknown distribution \(\mathcal{D}\). Let \(W \in \mathbb{R}^{n\times n}\) be a positive symmetric graph affinity matrix, with nodes corresponding to \(\mathcal{X}\), and let \(D\) be the corresponding diagonal degree matrix (i.e. \(D_{ii} = \sum_{j=1}^n W_{ij}\)). The Unnormalized Graph Laplacian is defined as \(L = D - W\). Other normalized Laplacian versions are the Symmetric Laplacian \(L_{\text{sym}} = D^{-\frac{1}{2}}LD^{-\frac{1}{2}}\) and the Random-Walk (RW) Laplacian \(L_{\text{rw}} = D^{-1}L\). \ourmethod\ is applicable to all of these Laplacian versions.
The eigenvalues of \(L\) can be sorted to satisfy \(0 = \lambda_1 \leq \lambda_2 \leq \cdots \leq \lambda_n\)
with corresponding eigenvectors \(v_1, \dots, v_n\) \citep{von2007tutorial}.
% It's important to note that for any Laplacian \(\lambda_1 = 0\) and \(v_1\) is the constant vector (i.e., \(v_1 = \frac{1}{n}\vec{1}\)).\\
It is important to note that the first pair (i.e., \(\lambda_1, v_1\)) is trivial - for every Laplacian matrix \(\lambda_1 = 0\), and for the unnormalized and RW Laplacians \(v_1 = \frac{1}{\sqrt{n}}\vec{1}\), namely the constant vector.

For a given target dimension \(k\), the first non-trivial \(k\) eigenvectors provide a natural non-linear low-dimensional embedding of the graph which is known as \textit{Spectral Embedding} (SE). In practice, we denote by \(V \in\mathbb{R}^{n\times k}\) the matrix containing the first non-trivial \(k\) eigenvectors of the Laplacian matrix as its columns (i.e., \(v_2, \dots, v_{k+1}\)). The SE representation of each sample \(x_i \in\mathbb{R}^d\) is the \(i\)th row of \(V\), i.e., \(y_i= (v_2(i),\dots, v_{k+1}(i))\).

\subsection{SpectralNet}
\label{sec:spectralnet}
A prominent method for addressing scalability and generalizability in Spectral Clustering (SC) is using deep neural networks, for example SpectralNet \citep{spectralnet}. SpectralNet follows a common methodology for transferring the problem of matrix decomposition to its smallest eigenvectors to an optimization problem, through minimization of the Rayleigh Quotient (RQ).

% as the smallest eigenvector of \(L\) minimizes the Rayleigh Quotient \(R_L(v) = v^TLv\), with its minimum value being the corresponding eigenvalue. In general, the Rayleigh Quotient (RQ) is defined as follows.
\begin{definition} \label{def:rq}
    The Rayleigh quotient (RQ) of a Laplacian matrix \(L \in \mathbb{R}^{n\times n}\) is a function \(R_L: \mathbb{R}^{n\times k} \to \mathbb{R}\) defined on \(A \in \mathbb{R}^{n\times k}\) by \[R_L(A) = \Tr(A^T L A),\]
\end{definition}
% Maybe V ... satisfies V = argmin_{A^T A = I} R(A)?

SpectralNet first minimizes the RQ on small batches, while enforcing orthogonality. Namely, it approximates \(\theta^*\) which minimizes 
\begin{equation}
    \mathcal{L}_{\text{spectralnet}}(\theta) = \frac{1}{m^2}R_L\big(f_{\theta}(X)\big) \quad s.t. \quad \frac{1}{m}f_{\theta}(X)^Tf_{\theta}(X) = I_{k\times k},
\end{equation}

where \(m\) is the minibatch size and \(X\in\mathbb{R}^{m\times d}\) contains the \(d\)-dimensional data samples as rows. Thereby, it learns a map \(f: \mathbb{R}^d \to \mathbb{R}^k\) (where \(d\) is the input dimension) which approximates the space spanned by the first \(k\) eigenfunctions of the Laplace-Beltrami operator on the underlying manifold \(\mathcal{D}\) \citep{belkin2006convergence, shi2015convergence}. Then, it clusters the representations via \(k\)-means. These eigenfunctions are a natural generalization of the SE to unseen points, enabling both scalable and generalizable spectral clustering.
% However, as it only approximates the aforementioned space, it is insufficient for approximating the SE.

% Our approach adopts the aforementioned optimization approach to SE, and suggests a way to solve this ambiguity. Note that this kind of ambiguity (i.e., rotation and reflection of the basis of the space) doesn't affect the results of the clustering.

\subsection{UMAP and Parametric UMAP}
\label{sec:pre_umap}

\begin{figure}
    \centering
    \includegraphics[width=0.7\linewidth]{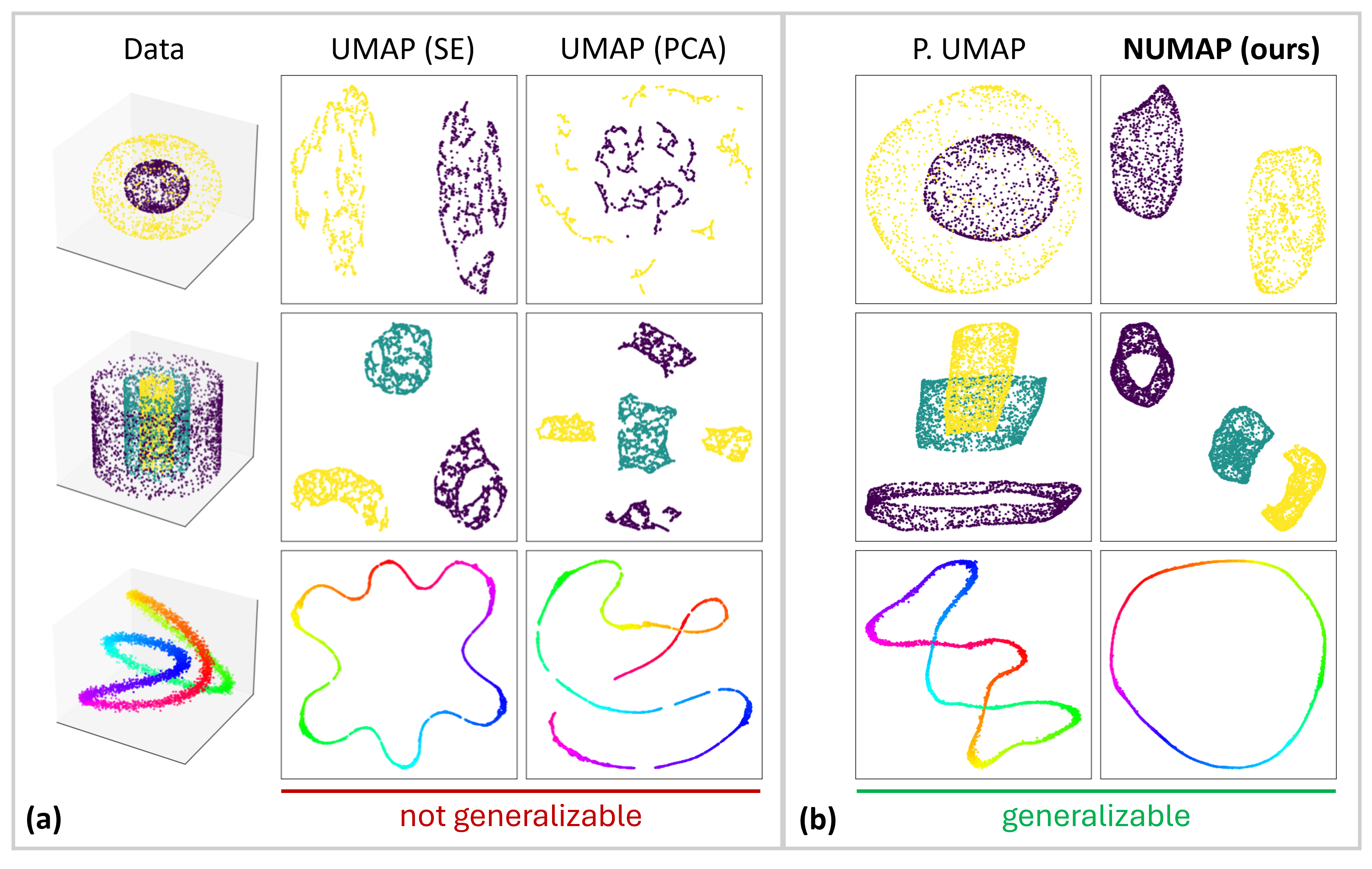}
    \caption{\textbf{Spectral Embedding is crucial for global structure preservation.} (a) Non-parametric UMAP preserves the global structure when initialized with SE, but fails to do so with PCA initialization - note the in-cluster separation in the middle row; (b) A similar effect is observed in generalizable implementations: P. UMAP, which does not involve SE, , fails to maintain global structure, resulting in overlapping clusters (e.g., first and second rows). In contrast, NUMAP (ours) achieves global structure preservation comparable to UMAP while supporting generalizability.}
    \label{fig:qualitative}
\end{figure}

UMAP \citep{umap} is a widely used visualization method, known for its scalability and its ability to preserve global structure. It constructs a graph from high-dimensional data and learns a low-dimensional representation by minimizing the KL-divergence between the input graph and the representation graph. The UMAP method can be divided into three components (summarized in Fig. \ref{fig:numap_method}): (1) constructing a graph which best captures the global structure of the input data; (2) initializing the representations via SE; (3) Learning the representations, via SGD, which best capture the original graph. This setup does not facilitate generalization, as both steps (2) and (3) lack generalizability.

Additionally, as discussed in \citep{kobak2021initialization}, UMAP primarily derives its global preservation abilities, as well as its consistency, from initializing the representations using SE. Therefore, the SE initialization serves as a critical step for UMAP to uphold the global structure (see demonstrations in Fig. \ref{fig:qualitative}a). Global preservation, in this context, refers to the separation of different classes, and avoiding the separation of existing classes. We refer the reader to \citep{kobak2021initialization} for a more comprehensive discussion about the effects of informative initialization on UMAP's performance.

Recently, a generalizable version of UMAP, known as Parametric UMAP (P. UMAP), was introduced \citep{sainburg2021parametric}. P. UMAP replaces step (3) with the training of a neural network. Importantly, it overlooks step (2), the SE initialization. Consequently, P. UMAP may struggle to preserve global structure, particularly when dealing with non-linear structures. Fig. \ref{fig:qualitative}b illustrates this phenomenon using several simple yet non-linear structures. These examples are particularly insightful for highlighting the importance of SE in preserving global structure, as the expected outcome of a good visualization is known. Noticeably, P. UMAP fails to preserve global structure (e.g., it does not separate different clusters).

% Good sentence: While most applications of UMAP involve projection from high-dimensional data, the projection from 3D serves as a useful analogy to understand how UMAP prioritizes global vs local structure depending on its parameters.

\section{Method}
\label{sec:method}
\subsection{Motivation}
It is well known that \(V\in\mathbb{R}^{n\times k}\), whose columns are the first \(k\) eigenvectors of \(L\) (i.e., those corresponding to the \(k\) smallest eigenvalues), minimizes \(R_L(A)\) under orthogonality constraint (i.e., \(A^TA=I\)) \citep{li2015rayleigh}.

However, a rotation and reflection ambiguity of the RQ prohibits a trivial adaptation of this concept to SE. Basic properties of trace imply that for any orthogonal matrix \(Q \in \mathbb{R}^{k\times k}\) the matrix \(U := VQ\) satisfies \(R_L(U) = R_L(V)\).
Thus, every such \(U\) also minimizes \(R_L\) under the orthogonality constraint, and therefore this kind of minimization solely is missing eigenvectors separation, which is crucial for many applications.

In fact, as stated in Lemma \ref{lemma:uniqueness}, the aforementioned form \(VQ\) is the only form of a minimizer of \(R_L\) under the orthogonality constraint. For conciseness, we provide our proof to the lemma in App. \ref{app:proof}.

\begin{lemma} \label{lemma:uniqueness}
    Every minimizer of \(R_L\) under the orthogonality constraint, is of the form \(VQ\), where \(V\) is the first \(k\) eigenvectors matrix of \(L\) and \(Q\) is an arbitrary squared orthogonal matrix.
\end{lemma}

An immediate result of Lemma \ref{lemma:uniqueness}
is that SpectralNet's method, using a deep neural network for RQ minimization (while enforcing orthogonality), does not lead to the SE. However, it only leads to the space spanned by the constant vector and the leading \(k-1\) eigenvectors of \(L\), with different rotations and reflections for each run. Therefore, each time the RQ is minimized, it results in a different linear combination of the smallest eigenvectors. Although this is sufficient for clustering purposes, as we search for reproducibility, consistency, and separation of the eigenvectors, the RQ cannot solely provide the SE, necessitating the development of new techniques in \ourmethod.

\subsection{\ourmethod}

\paragraph{Setup.} Here we present the two key components of \ourmethod, a scalable and generalizable SE method. We consider the following setup: Given a training set \(\mathcal{X} \subseteq \mathbb{R}^d\) and a target dimension \(k\), we construct an affinity matrix \(W\), and compute an approximation of the leading eigenvectors of its corresponding Laplacian. In practice, we first utilize SpectralNet \citep{spectralnet} to approximate the space spanned by the first \(k + 1\) eigenfunctions of the corresponding Laplace-Beltrami operator, and then find each of the \(k\) leading eigenfunctions within this space (i.e. the SE). Namely, \ourmethod\ computes a map \(F_{\theta}: \mathbb{R}^d \to \mathbb{R}^k\), which approximates the map \(\bar{f} = (f_2, \dots, f_{k+1})\), where \(f_i\) is the \(i\)th eigenfunction of the Laplace-Beltrami operator on the underlying manifold \(\mathcal{D}\).

\begin{figure}[]
    \centering
    \begin{subfigure}[b]{0.56\textwidth}
        \centering
        \includegraphics[width=\textwidth]{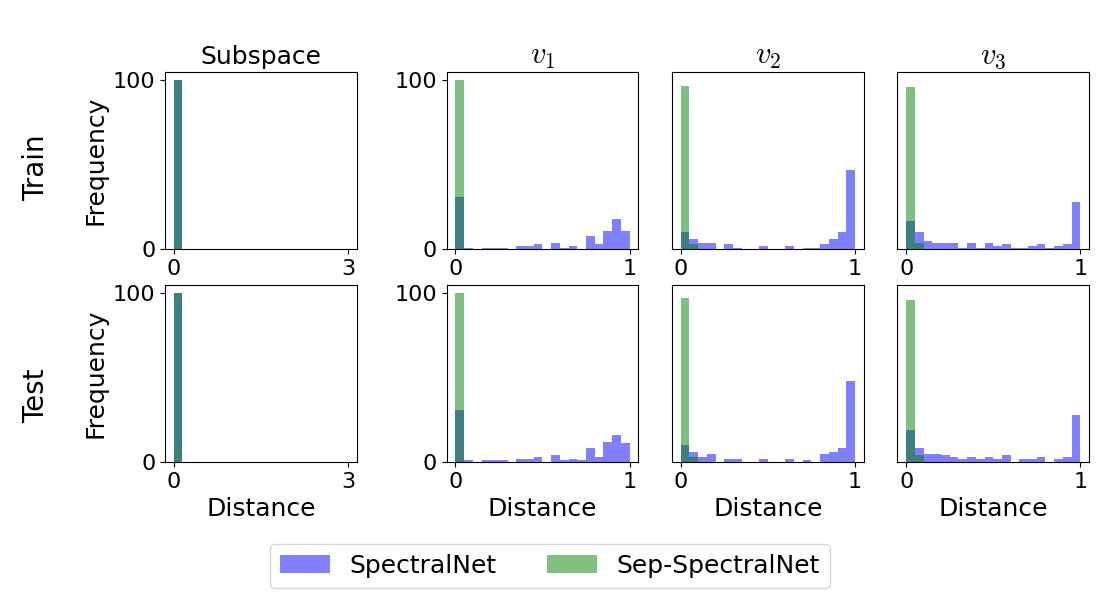}
        \caption{}
        \label{fig:moon_hists}
    \end{subfigure}
    \begin{subfigure}[b]{0.43\textwidth}
        \centering
        \includegraphics[width=\textwidth]{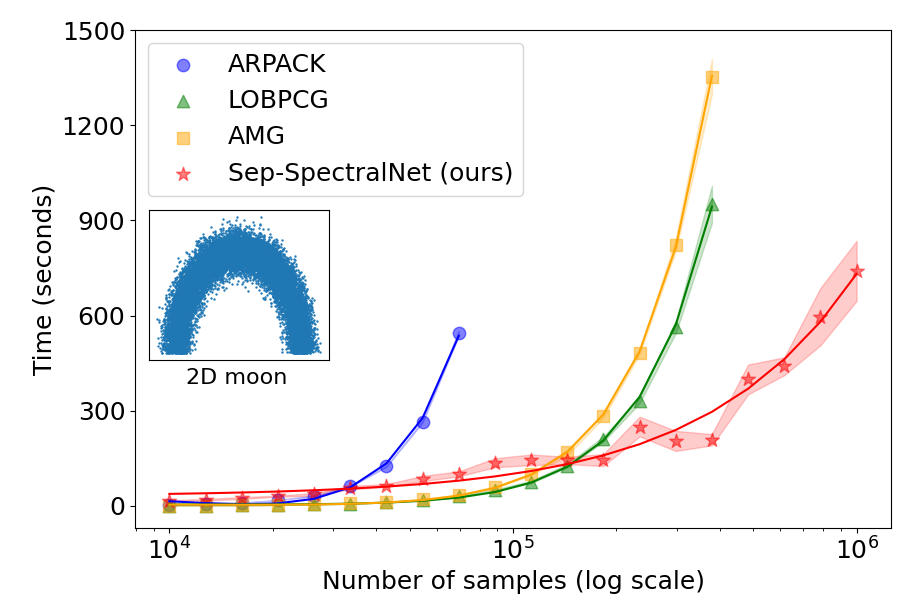}
        \caption{}
        \label{fig:scalability}
    \end{subfigure}
    \caption{\textbf{(a) \ourmethod\ separates the eigenvectors.} Approximation of the 2-dimensional SE of the moon dataset (a 2D moon linearly embedded into 10D input space) using SpectralNet (in blue) and \ourmethod\ (in green) over 100 runs, on train (top row) and test (bottom row). Left column: distribution of the Grassmann distance between the output and true subspace. Second to Fourth columns: distribution of the \(\text{sin}^2\) distance between each output and true eigenvector separately. Evidently, while SpectralNet's errors are distributed over a large range of values, \ourmethod's errors are small, capturing only the smallest error bin in the figure, indicating on eigenvector separation. \textbf{(b) \ourmethod's training is as scalable non-generalizable spectral methods.} Running times of SE using \ourmethod\ vs. other methods on the moon dataset, relative to the number of samples, and with standard deviation confidence intervals. Evidently, \ourmethod\ is the fastest asymptotically.}
\end{figure}

\paragraph{Eigenspace approximation.} As empirically showed in \citep{spectralnet}, and motivated from Lemma \ref{lemma:uniqueness}, SpectralNet loss is minimized when \(F_{\theta} = T \circ (f_1, \dots, f_{k+1})\), where \(T:\mathbb{R}^{k+1} \to \mathbb{R}^{k+1}\) is an arbitrary isometry. That is, \(F_{\theta}\) approximates the space spanned by the first \(k+1\) eigenfunctions. However, the SE (i.e. each of the leading eigenfunctions) is poorly approximated. Each time the RQ is minimized, the eigenfunctions are approximated up to a different isometry \(T\). Fig. \ref{fig:moon_hists} demonstrates this phenomenon on the toy moon dataset -  a noisy half circle linearly embedded into 10-dimension input space (see Sec. \ref{sec:GrEASE_exps}). Employing SpectralNet approach indeed enables us to consistently achieve a perfect approximation of the space (i.e., the errors at the left histograms are accumulated around 0). However, when comparing vector to vector, it becomes apparent that the SE was seldom attained.

\paragraph{SE approximation.} To separate the eigenvectors, and thereby consistently get the SE, we present a simple use of Lemma \ref{lemma:uniqueness}. Notice that based on Lemma \ref{lemma:uniqueness} we can compute a rotated version of the diagonal eigenvalues matrix. Namely, 
\begin{equation*}
    (VQ)^T L (VQ) = Q^T V^T L V Q = Q^T \Lambda Q =: \Tilde{\Lambda},
\end{equation*}
where \(\Lambda\) is the diagonal eigenvalues matrix. Due to the uniqueness of eigendecomposition, the eigenvectors and eigenvalues of the small matrix \(\Tilde{\Lambda} \in \mathbb{R}^{k+1\times k+1}\) are \(Q^T\) and \(\text{diag}(\Lambda)\), respectively. Hence, by diagonalizing \(\Tilde{\Lambda}\) we get the eigenvalues and are also able to separate the eigenvectors (i.e., approximate the SE).

In practice, as \(Q\) is a property of SpectralNet optimization (manifested by the parameters), we compute the matrix \(\Tilde{\Lambda}\) by averaging over a few random minibatches and diagonalize it. Thereby, making this addition very cheap computationally. The eigenvectors matrix of \(\Tilde{\Lambda}\) is the inverse of the orthogonal matrix \(Q\), and hence by multiplying the output of the learned map \(F_{\theta}\) by this matrix, the SE is retained. Also, the eigenvalues of \(\Tilde{\Lambda}\) are the eigenvalues of \(L\).

This step is simple and elegant, yet its impact is significant. The effect of this intentional rotation is represented in the Fig. \ref{fig:moon_hists}. \ourmethod\ is not only able to consistently approximate the space, but also to approximate each eigenvector.

\paragraph{Algorithms Layout.} Our end-to-end training approach is summarized in Algorithms \ref{alg:spectralnet} and \ref{alg:GrEASE} in App. \ref{app:algo}. We first train \(F_{\theta}\) to approximate the first eigenfunctions up to isometry (Algorithm \ref{alg:spectralnet}) \citep{spectralnet}, then compute \(Q^T\) and \(\Lambda\) to separate the eigenvectors and recover the SE and corresponding eigenvalues (Algorithm \ref{alg:GrEASE}). App. \ref{app:additional_considerations} details additional considerations about the implementation.

Once we have \(F_{\theta}\) and \(Q^T\), computing the embeddings of the train set or of new test points (i.e., out-of-sample extension) is straightforward: we simply propagate each test point \(x_i\) through the network \(F_{\theta}\) to obtain their embeddings \(\Tilde{y_i}\), and use \(Q^T\) to get the SE embeddings \(y_i = \Tilde{y_i}Q^T\).

\paragraph{Time and Space complexity.} As the network iterates over small batches, and the post-processing operation is much cheaper, \ourmethod's time complexity is approximately linear in the number of samples. This is also demonstrated in Fig. \ref{fig:scalability}, where the continuous red line, representing linear regression, aligns with our empirical results. App. \ref{app:additional_considerations} provides a discussion about the complexity of \ourmethod. \ourmethod\ is also highly memory-efficient, requiring only minibatch-sized graphs or matrices in memory, rather than the full graph.

\subsection{NUMAP}
Here, we demonstrate \ourmethod's extension to UMAP, one of many methods that may benefit from a generalizable SE. As discussed in Sec. \ref{sec:pre_umap}, the SE initialization is crucial for the global preservation abilities of UMAP. Therefore, we seek a method to incorporate SE into a generalizable version of UMAP. A na\"ive approach would be to fine-tune \ourmethod\ using UMAP loss (Eq. \ref{eq:umap_loss}). However, during this implementation, we encountered the phenomenon of catastrophic forgetting (see App. \ref{app:numap_ft}).

\begin{figure}
    \centering
    \includegraphics[width=\linewidth]{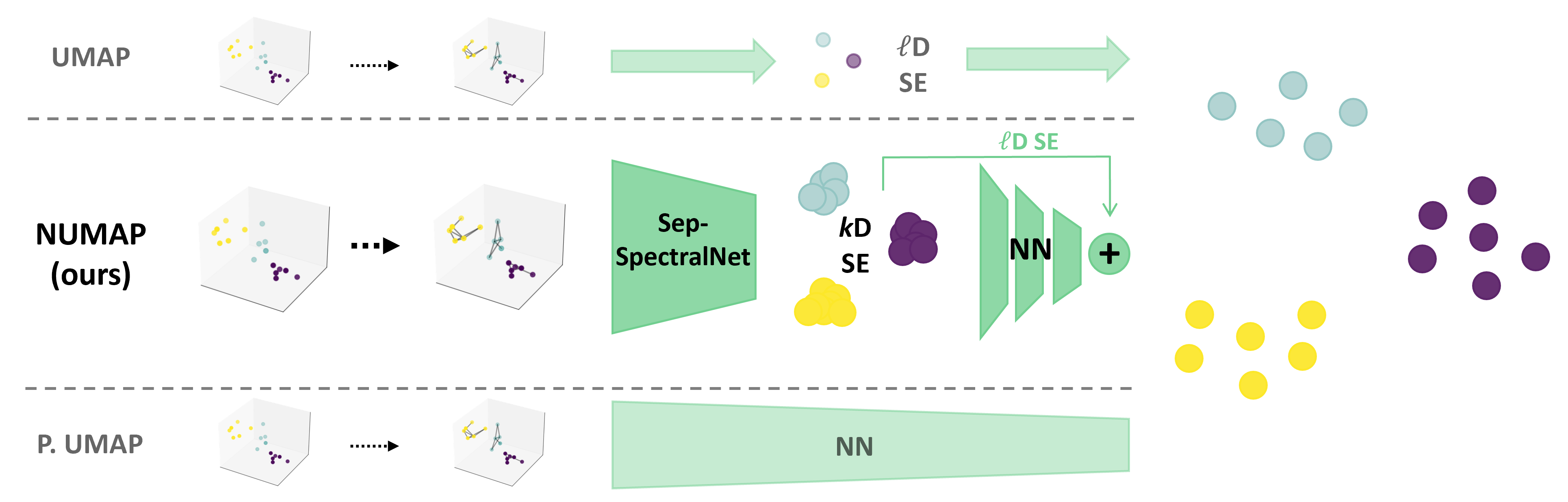}
    \caption{\textbf{Incorporating \ourmethod\ for generalizable UMAP.} UMAP vs. NUMAP vs. P. UMAP overview. NUMAP integrates SE, as in UMAP, while enabling generalization. First, it learns a parametric representation of the \(k\)-dimensional SE. Then, it learns a map from the SE to the final \(\ell\)-dimensional space, incorporating residual connections from the first \(\ell\) input components to the output layer.}
    \label{fig:numap_method}
\end{figure}

The core of our idea is illustrated in Fig. \ref{fig:numap_method}. Initially, we use \ourmethod\ to learn a parametric representation of the \(k\)-dimensional SE of the input data. Subsequently, we train an NN to map from the SE to the final \(\ell\)-dimensional embedding space (usually for \(\ell = 2, 3\)), utilizing UMAP contrastive loss. The objective of the second NN is to identify representations that best capture the local structure of the input data graph. SE transforms complex non-linear structures into simpler linear structures, allowing the second NN to preserve both local and global structures effectively. To enhance this capability, we incorporate residual connections in the second NN, \(G_{\theta}\), which link the first \(\ell\) components of the input (the \(\text{SE}_{1:\ell}\)) directly to the output. Thus, the NUMAP embedding is given by \(Y = \text{SE}(X)_{1:\ell} + G_{\theta}(X)\), that minimizes the UMAP's loss
\begin{equation} \label{eq:umap_loss}
\sum_{e \in E} w_h(e) \log \left( \frac{w_h(e)}{w_l(e)} \right) + (1 - w_h(e)) \log \left( \frac{1 - w_h(e)}{1 - w_l(e)} \right),    
\end{equation}

where \(w_h(e), w_l(e)\) are the corresponding weights of the edge \(e\) in the high-dimensional input (\(X\)) graph and the low-dimensional output (\(Y\)) graph, respectively. It should be noted that this could not have been possible without \ourmethod's ability to separate the eigenvectors, as the first \(\ell\) components of SpectralNet's output are merely arbitrary linear combinations of the \(k\)-dimensional SE. Also, it would not be practical without \ourmethod's inherent generalizability and scalability. Fig. \ref{fig:qualitative} demonstrates this capability with several simple structures.

% \begin{table}[t]
% \scriptsize
% \caption{A comparison between NUMAP, P. UMAP and UMAP visualization on real-world datasets. The values are the mean and standart deviation of the measures, over 10 runs. The arrows indicates lower/higher is better.}
% \label{tab:NUMAP_results}
% \begin{center}

% \begin{tabularx}{\textwidth}{C|C|C|C}
% & Step 1 & Step 2 & Step 3 \\
% \hline
% UMAP & Compute a graphical representation of the dataset
%  & Initialize the embeddings with \textbf{2D SE}
%  & Learn an embedding that best preserves the graph local structure \\
% \hline
% P. UMAP & Compute a graphical representation of the dataset
%  & ---- & Learn \textbf{a network from the input space} that best preserves the graph local structure \\
% \hline
% NUMAP (ours) & Compute a graphical representation of the dataset
%  & learn a \textbf{parametric} approximation of the \textbf{kD SE} of the graph
%  & Learn \textbf{a network from the SE} that best preserves the graph local structure \\
% \end{tabularx}
% \end{center}
% \end{table}

\subsection{Additional Applications}
In this section we seek to highlight \ourmethod's potential impact on important tasks and applications (besides generalizable UMAP), as it integrates generalizability, scalability and eigenvectors separation.
As discussed in Sec. \ref{sec:intro}, SE is applied across various domains, many of which can benefit generalizability capabilities by replacing the current SE implementation with \ourmethod. We therefore elaborate herein the significance of SE in selected applications, and discuss how \ourmethod, as a generalizable approximation of it, can enhance their effectiveness and applicability.

\paragraph{Fiedler vector and value.} A special case of SE is the Fiedler vector and value \citep{fiedler1973algebraic, fiedler1975property}. The Fiedler value, also known as algebraic connectivity, refers to the second eigenvalue of the Laplacian matrix, while the Fiedler vector refers to the associated eigenvector. This value quantifies the connectivity of a graph, increasing as the graph becomes more connected. Specifically, if a graph is not connected, its Fiedler value is 0. The Fiedler vector and value are a main topic of many works \citep{andersen2006local, barnard1993spectral, kundu2004automatic, shepherd2007amino, cai2018dynamic, zhu2021fiedler, tam2020fiedler}.
As \ourmethod\ is able to distinguish between the eigenvectors and approximate the eigenvalues, it has the capability to approximate both the Fiedler vector and value, while also generalizing the vector to unseen samples (see Sec. \ref{sec:GrEASE_exps}).

\paragraph{Diffusion Maps.} A popular method which incorporates SE, alongside the eigenvalues of the Laplacian matrix, is Diffusion Maps \citep{diffusionMaps}. Diffusion Maps embeds a graph (or a manifold) into a space where the pairwise Euclidean distances are equivalent to the pairwise Diffusion distances on the graph. This approach is widely used (e.g., \citep{ghamari2024sampling, hardin2024revealing, park2021differences}).

In practice, for an \(k\)-dimensional embedding space and a given \(t \in \mathbb{N}\), Diffusion Maps maps the points to the leading eigenvectors of the RW-Laplacian matrix of the data as follows: \[X \to \begin{pmatrix}
    (1-\lambda_2)^t v_2 & \cdots & (1-\lambda_{k+1})^t v_{k+1}
    \end{pmatrix} = Y,\]
where \(X \in \mathbb{R}^{n\times d}\) is a matrix containing each input point as a row, and \(Y \in \mathbb{R}^{n \times k}\) is a matrix containing each of the representations as a row. As \ourmethod\ is able to approximate both the eigenvectors and eigenvalues of the Laplacian matrix, it is able to make Diffusion Maps generalizable and efficient (Sec. \ref{sec:GrEASE_exps}).

\subsection{Evaluating UMAP embedding - Grassmann Score}
\label{subsec:grassmann_score}
Common evaluation methods for dimensionality reduction, particularly for visualization, are predominantly focused on local structures. For instance, \citet{umap, kawase2022parametric} use kNN accuracy and Trustworthiness, which only account for the local neighborhoods of each point while overlooking global structures such as cluster separation. One global evaluation method is the Silhouette score, which measures the clustering quality of the classes within the embedding space. However, this score does not capture the preservation of the overall global structure.

To address this gap, we propose a new evaluation method, specifically appropriate for assessing global structure preservation in graph-based dimensionality reduction methods (e.g., UMAP, t-SNE). The leading eigenvectors of the Laplacian matrix are known to encode crucial global information about the graph \citep{laplacianEigenmaps}. Thus, we measure the distance between the global structures of the original and embedding manifolds using the Grassmann distance between the first eigenvectors of their respective Laplacian matrices. We refer to this method as the \textit{Grassmann Score} (GS). 

\begin{figure}
    \centering
    \includegraphics[width=0.9\linewidth]{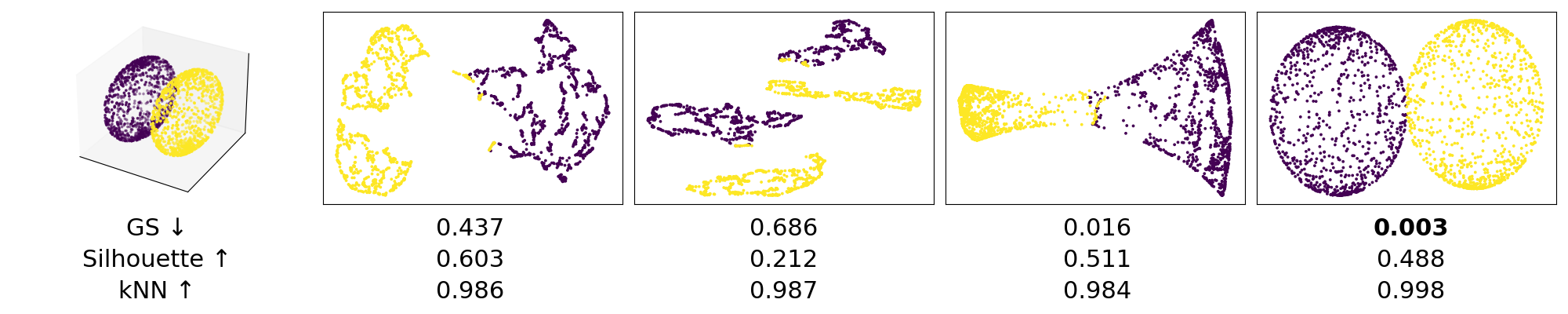}
    \caption{\textbf{Grassmann Score (GS) reflects global structure preservation.} A demonstration of the alignment between intuitive expectations and GS on a toy dataset of two 3D tangent spheres. Four possible 2D embeddings of the dataset are shown, alongside their corresponding GS, kNN accuracy, and Silhouette score. The rightmost visualization best preserves the global structure, forming two tangent circles in 2D. Notably, kNN fails to distinguish between the embeddings, while Silhouette favors the most clustered one, despite its poor reflection of the original structure. In contrast, GS successfully captures the preservation of global structure, assigning the highest score to the most faithful embedding.}
    \label{fig:grassmann_score}
\end{figure}

It is important to note that GS includes a hyper-parameter - the number of eigenvectors considered. Increasing the number of eigenvectors incorporates more local structure into the evaluation. A natural choice for this hyperparameter is 2, which corresponds to comparing the Fiedler vectors (i.e., the second eigenvectors of the Laplacian). The Fiedler vector is well known for encapsulating the global information of a graph \citep{fiedler1973algebraic, fiedler1975property}. Accordingly, for simplicity and unless specified otherwise, the GS is computed using the first two eigenvectors. Fig. \ref{fig:grassmann_score} demonstrates GS (alongside Silhouette and kNN scores for comparison) on a few embeddings of two tangent spheres, independently to the embedding methods. Notably, the embedding on the right appears to best preserve the global structure, as indicated by the smallest GS value. In contrast, the kNN scores are comparable across all embeddings (e.g., kNN ignores separation of an existing class), and the Silhouette score even favors other embeddings. In App. \ref{app:grassmann_score} we mathematically formalize GS and provide additional examples of embeddings and their corresponding GS. These examples further support the intuition that GS effectively captures global structure preservation better than previous measures.

\section{Experiments}
\label{sec:exps}

\subsection{Eigenvectors Separation - Generalizable SE}
\label{sec:GrEASE_exps}
In this section, we demonstrate \ourmethod's ability to approximate and generalize the SE using four real-world datasets: CIFAR10 (via their CLIP embedding); Appliances Energy Prediction dataset \citep{misc_appliances_energy_prediction_374_dataset}; Kuzushiji-MNIST (KMNIST) dataset \citep{clanuwat2018deep}; Parkinsons Telemonitoring dataset \citep{misc_parkinsons_telemonitoring_189_dataset}. Particularly, we compare our results with SpectralNet, which has been empirically shown to approximate the SE space. However, as our results demonstrate, SpectralNet is insufficient for accurately approximating SE. For additional technical details regarding the datasets, architectures and training procedures, we refer the reader to Appendix \ref{app:technical_details}.

\paragraph{Evaluation Metrics.} To assess the approximation of each eigenvector (i.e., the SE), we compute the \(\text{sin}^2\) of the angle between each predicted and ground truth vector. This can be viewed as the 1-dimensional case of the Grassmann distance, a well-known metric for comparing equidimensional linear subspaces (see formalization in App. \ref{app:grassmann_score}). Concerning the eigenvalues approximation evaluation, we measure the Pearson Correlation between the predicted and true eigenvalues (computed via SVD).

% \begin{figure}
%     \centering
%     \includegraphics[width=0.8\linewidth]{figures/approx_barplot.png}
%     \caption{A comparison between GrEASE and SpectralNet SE and Fiedler Vector (FV) approximation on real-world datasets. The values are the mean and standart deviation of the \(\text{sin}^2\) distance between the predicted and true eigenvector of the test set, over 10 runs. Lower is better. GrEASE ability to separate the eigenvectors is evident.}
%     \label{fig:GrEASE_results}
% \end{figure}

\begin{figure}[t]
    \centering
    \includegraphics[width=0.95\linewidth]{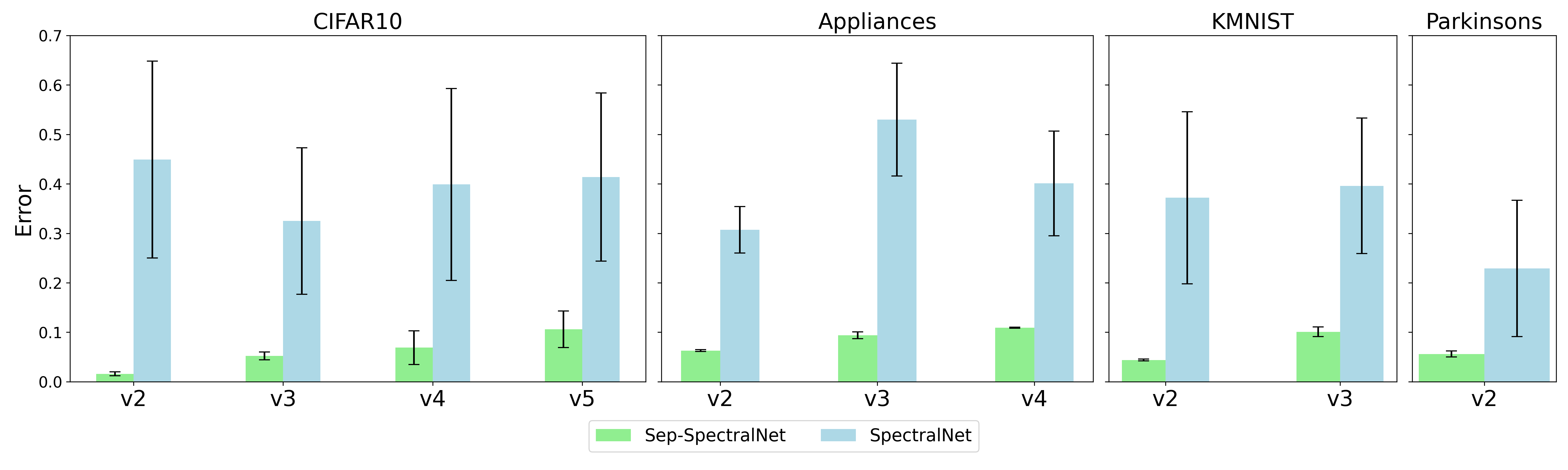}
    \caption{\textbf{\ourmethod\ successfully separates the eigenvectors.} A comparison between \ourmethod\ and SpectralNet of SE and Fiedler Vector (v2) approximation on real-world datasets. The values are the mean and standard deviation of the \(\text{sin}^2\) distance between the predicted and true eigenvector of the test set, over 10 runs. Lower is better. Notably, \ourmethod's errors are significantly smaller than SpectralNet's, supporting \ourmethod's ability to separate the eigenvectors.}
    \label{fig:GrEASE_results}
\end{figure}

Fig. \ref{fig:GrEASE_results} presents our results on the real-world datasets. \ourmethod’s output is used directly, while SpectralNet’s predicted eigenvectors are resorted to minimize the mean \(\text{sin}^2\) distance. The results clearly show that \ourmethod\ consistently produces significantly more accurate SE approximations compared to SpectralNet, due to the improved separation of the eigenvectors.

Additionaly, note the \ourmethod\ approximates the eigenvalues as well. When concerning a series of Laplacian eigenvalues, the most important property is the relative increase of the eigenvalues \citep{diffusionMaps}. \ourmethod\ demonstrates a strong ability to approximate this property. To see this, we repeated \ourmethod's eigenvalue approximation (10 times) and calculated the Pearson correlation between the predicted and accurate eigenvalues vector. We compared the first 10 eigenvalues.
% (see Fig. \ref{fig:moon_eigvals} for illustration of the moon eigenvalues prediction).
The resulting mean correlation and standard deviation are: Parkinsons Telemonitoring: \(\textbf{0.917}_{\pm 0.0381}\); Appliances Energy Prediction: \(\textbf{0.839}_{\pm 0.0342}\);

\subsection{Scalability}
Noteworthy, \ourmethod\ not only generalizes effectively but also does so more quickly than the most scalable (yet non-generalizable) existing methods. Fig. \ref{fig:scalability} demonstrates this point on the toy moon dataset - a 2D moon linearly embedded into 10D input space. To evaluate scalability, we measured the computation time required for SE approximation, for an increasing number of samples. We compared the results with the three most popular methods for sparse matrix decomposition, which are currently the fastest implementations: ARPACK \citep{lehoucq1998arpack}, LOBPCG \citep{benner2011locallyLOBPCG}, and AMG \citep{brandt1984algebraicAMG}. For each number of samples, we calculated the Laplacian matrix that is 99\% sparse. Each method was executed five times, initialized with different seeds. Notably, for higher numbers of samples, \ourmethod\ converges significantly faster. Notably, \ourmethod maintains the same scalability as SpectralNet, since the separation step incurs negligible overhead compared to the overall training time.

\subsection{NUMAP - generalizable UMAP}
In this section, we demonstrate NUMAP's ability to preserve global structure, while enabling fast inference of test points, and it's ability to enable time-series UMAP visualization. We compare our results with P. UMAP, with the target dimensionality set to 2. We refer to App. \ref{app:numap_ft} for various ablations.

We consider four real-world datasets: CIFAR10 (via their CLIP embedding); Appliances Energy Prediction dataset; Wine \citep{wine_109}; Banknote Authentication \citep{banknote_authentication_267}. For additional technical details regarding the datasets, architectures and training procedures, we refer to App. \ref{app:technical_details}.

\paragraph{Evaluation Metrics.} To evaluate and compare the embeddings, we employed both local and global evaluation metrics. For local evaluation, we used the well-established accuracy of a kNN classifier on the embeddings \citep{umap, sainburg2021parametric}, which is applicable only on classed data.
For global evaluation, we use GS (see discussion in Sec. \ref{subsec:grassmann_score}).

\paragraph{Local Structure.} Tab. \ref{tab:NUMAP_local_results} presents the kNN results on the real-world datasets. These results are comparable across the methods. Namely, NUMAP does not compromise on local structure.

\begin{wrapfigure}{r}{0.4\textwidth}
    \centering
    \includegraphics[width=0.38\textwidth]{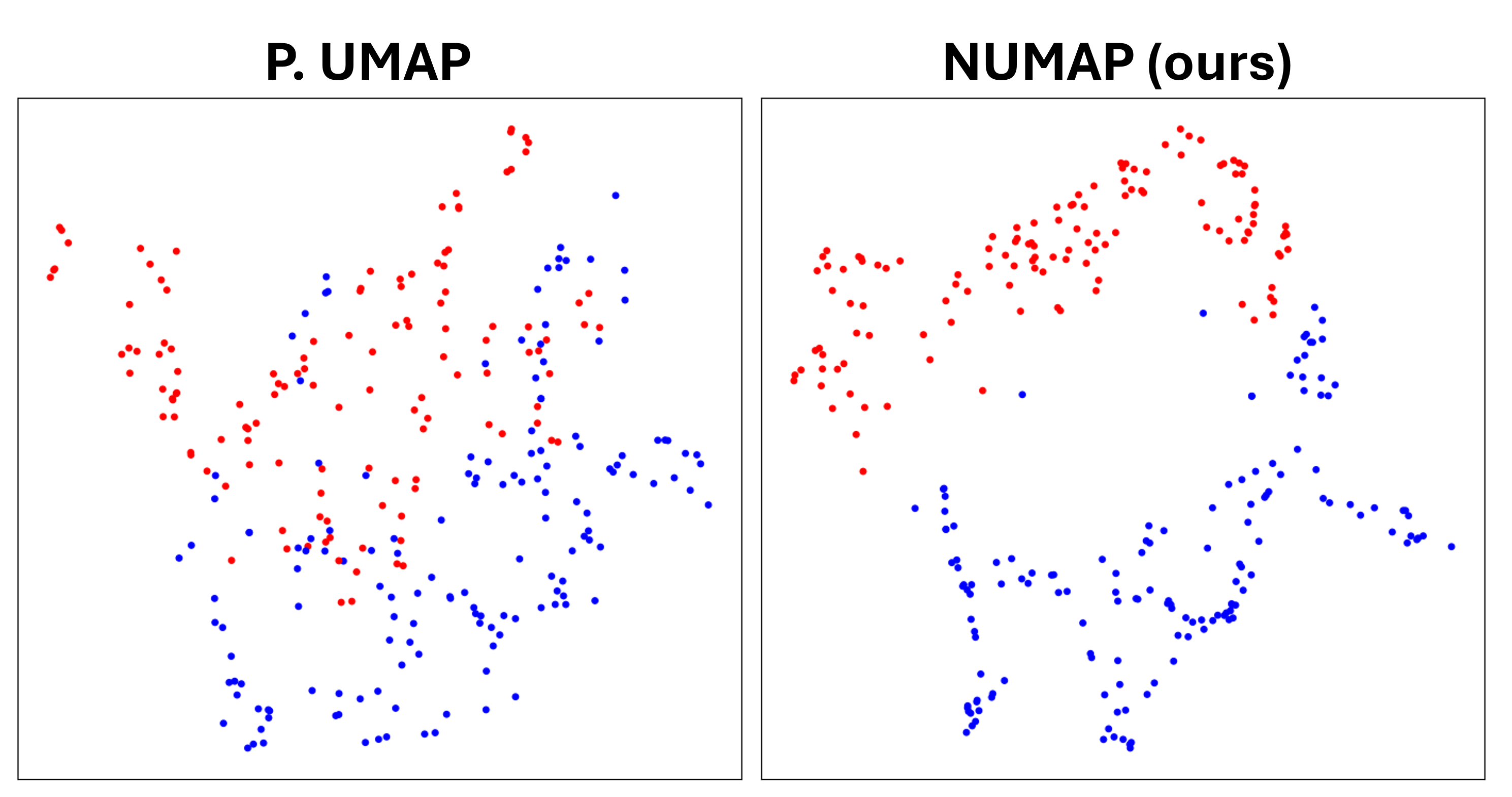}
    \caption{\textbf{NUMAP and P. UMAP visualization of the Banknote dataset test set.} NUMAP’s enhanced ability to preserve global structure is evident in the clearer separation between clusters compared to P. UMAP.}
    \label{fig:banknote_visus}
\end{wrapfigure}

\paragraph{Global Structure.} Tab. \ref{tab:NUMAP_global_results} presents the GS results on these datasets. The results underscore that NUMAP better captures the global structure. Namely, NUMAP enhances global structure preservation while not compromising local structure preservation.

In Fig. \ref{fig:qualitative} and Fig. \ref{fig:banknote_visus}, we supplement the empirical results with qualitative examples. Fig. \ref{fig:qualitative} presents simple non-linear synthetic 3-dimensional structures and their 2-dimensional visualizations using UMAP (non-parametric), P. UMAP and NUMAP. These examples are particularly insightful, as the expected outcome of a good visualization is known. UMAP (using its default configuration, SE initialization) accurately preserves the global structure in its 2-dimensional representations, but lack the ability to generalize to unseen points. Among the generalizable methods (Fig. \ref{fig:qualitative}b), P. UMAP fails to preserve the global structure: in the top two rows, it does not separate the clusters, while in the bottom row, it introduces undesired color overlaps. In contrast, NUMAP effectively preserves these separations and avoids unnecessary overlapping.
Fig. \ref{fig:banknote_visus} further demonstrates NUMAP's ability to preserve global structure, as evidenced by the improved class separation in the Banknote dataset.

\begin{table}[t]
\footnotesize
\begin{center}
\begin{tabular}{llllllll}
\multirow{1}{*}{\bf Metric} & \multirow{1}{*}{\bf Method} & Cifar10 & Appliances & Wine & Banknote & \\
\hline\\

\multirow{2}{*}{kNN \(\uparrow\)} & P. UMAP & {0.908\(_{\pm0.004}\)} & - & {0.953\(_{\pm0.033}\)} & {0.927\(_{\pm0.023}\)} \\

& \bf NUMAP (ours) & {0.905\(_{\pm0.004}\)} & - & {0.950\(_{\pm0.024}\)} & {0.986\(_{\pm0.004}\)} \\

\hline
\end{tabular}
\end{center}
\caption{\textbf{NUMAP local structure preservation is comparable with P. UMAP.} kNN accuracy results of P. UMAP and NUMAP on labeled real-world datasets (Appliances dataset is excluded due to the absence of labels). The values are the mean and standard deviation measures on the test set, over 10 runs.}
\label{tab:NUMAP_local_results}
\end{table}

\begin{table}[t]
\footnotesize
\begin{center}
\begin{tabular}{llllllll}
\multirow{1}{*}{\bf Metric} & \multirow{1}{*}{\bf Method} & Cifar10 & Appliances & Wine & Banknote & \\
\hline\\

\multirow{2}{*}{GS \(\downarrow\)} & P. UMAP & {0.133\(_{\pm0.069}\)} & {0.710\(_{\pm0.293}\)} & {0.549\(_{\pm0.180}\)} & {0.722\(_{\pm0.079}\)} \\

& \bf NUMAP (ours) & \bf {0.054\(_{\pm0.021}\)} & \bf {0.261\(_{\pm0.020}\)} & \bf {0.429\(_{\pm0.124}\)} & \bf {0.618\(_{\pm0.131}\)} \\
\hline

& p-value & 0.004 & 0.0002 & 0.118 & 0.056 \\
\hline

\end{tabular}
\end{center}
\caption{\textbf{NUMAP better preserves global structure.} P. UMAP and NUMAP Grassmann Score (GS) results on real-world datasets. The values are the mean and standard deviation measures on the test set, over 10 runs. The best in mean is highlighted. Notably, NUMAP achieves better global structure preservation.}
\label{tab:NUMAP_global_results}
\end{table}

\begin{figure}
    \centering
    \includegraphics[width=\linewidth]{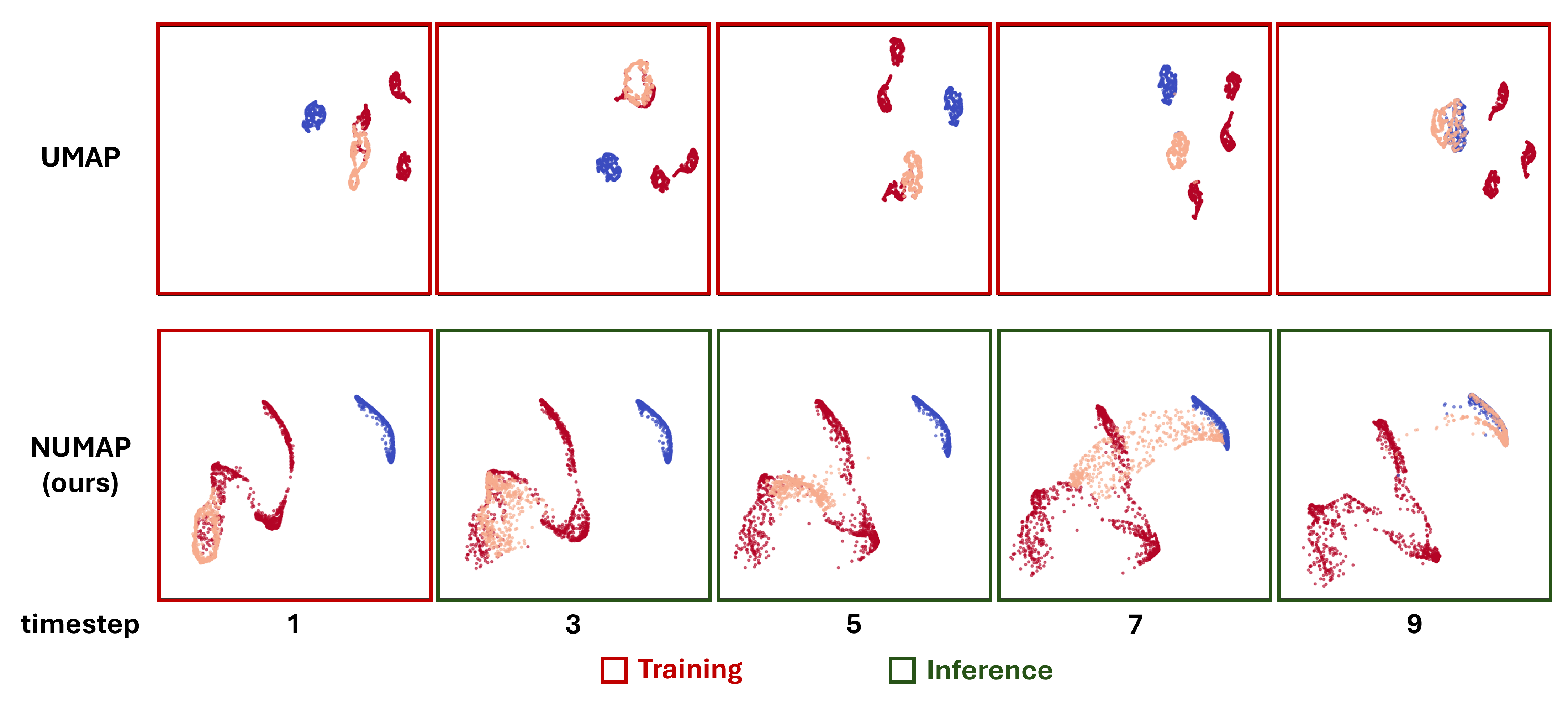}
    \caption{\textbf{Time-series data visualization using NUMAP.} A comparison between UMAP and NUMAP visualizations of a dynamical system. In the UMAP visualization, the structure and cluster positions vary significantly across timesteps, hindering the interpretability of the underlying process. In contrast, NUMAP produces consistent structures over time, clearly revealing the transition of the pink points between the red and blue regions. Notably, NUMAP requires training only on the first two timesteps, enabling efficient generalization to the rest of the sequence.}
    \label{fig:numap_teimesteps}
\end{figure}

\paragraph{Time-series data visualization.} Fig. \ref{fig:numap_teimesteps} shows a simulation time-series data, which can be viewed as a simulation of cellular differentiation. Specifically, we may consider differentiation of hematopoietic stem cells (also known as blood stem cells), which are known to differentiate into many types of blood cells, to T-cells. The process involves two kinds of cells (represented by their gene expressions; red and blue samples in the figure). One represents stem cells, while the other T-cells. A group of cells (colored in pink in the figure) then gradually transitions from stem cells to T-cells. At the top row we use UMAP to visualize each time step, while at the bottom we train NUMAP on the first two time-steps and only inference the rest. UMAP is inconsistent over time-steps, which makes it impractical for understanding change and progression. It also has to train the embeddings at each time-step separately. In contrast, NUMAP only trains on the first two time-steps, and the embeddings of the later time-steps are immediate from inference. This also enables consistency over time, and makes the trend and process visible and understandable.

\section{Conclusions}
We introduced \ourmethod, a deep-learning approach for approximate SE. \ourmethod\ addresses the three primary drawbacks of current SE implementation: generalizability, scalability and eigenvectors separation. By incorporating a post-processing diagonalization step, \ourmethod\ enables eigenvectors separation without compromising SpectralNet's generalizability or scalability. Remarkably, this one-shot post-processing operation lays the groundwork for a wide range of new applications of SE, which would not have been possible without its scalable and generalizable implementation. It also presents a promising pathway to enhance current applications of SE.

In particular, we also presented NUMAP, a novel extension of \ourmethod\ for generalizable UMAP visualization. We believe the integration of generalizable SE with deep learning can have a significant impact on unsupervised learning methods. However, exploring alternative inputs to the UMAP loss network is an interesting direction for future work. Finally, further research should delve into exploring the applications of SE across various fields.

\bibliography{main_revised}
\bibliographystyle{tmlr}

\appendix
\section{Proof of Lemma \ref{lemma:uniqueness}}\label{app:proof}
First, we remind an important property of the Rayleigh Quotient.
\begin{remark} \label{remark:rayleigh}
    The Rayleigh Quotient of a positive semi-definite matrix \(L \in \mathbb{R}^{n\times n}\) with eigenvectors \(v_1, \dots, v_n\) corresponding to the eigenvalues \(\lambda_1 \leq \cdots \leq \lambda_n\), \(R_L\) satisfies \(arg\min_{||v||=1} R_L(v) = v_1\) and for each \(i > 1\) \(arg\min_{||v|| = 1} R_L(v) = v_i\) for \(v \perp v_1, \dots, v_{i-1}\) \text{\citep{li2015rayleigh}}.
\end{remark}
\textbf{Lemma \ref{lemma:uniqueness}.} Let \(L \in \mathbb{R}^{n\times n}\) be an Unnormalized Laplacian matrix and \(R_L:O(n, k) \to \mathbb{R}\) its corresponding RQ, and Let \(A\) be a minimizer of \(R_L\). Denote \(V \in \mathbb{R}^{n\times k}\) as the matrix containing the first \(k\) eigenvectors of \(L\) as its columns, and \(\Lambda\) the corresponding diagonal eigenvalues matrix. Then, there exists an orthogonal matrix \(Q \in \mathbb{R}^{k\times k}\) such that \(A = VQ\).
\begin{proof}
    As \(V\) minimizes \(R_L\), we get that \(\min_U R_L(U) = R_L(V) = \sum_{i=1}^k \lambda_i\), where \(0 = \lambda_1 \leq \lambda_2 \leq \cdots \leq \lambda_n\) are the eigenvalues of \(L\). This yields \[R_L(A) = \Tr(A^TLA) = \sum_{i=1}^k \lambda_i\]
    \(A^TLA\) is symmetric, and hence orthogonally diagonalizable, which means there exists an orthogonal matrix \(Q \in \mathbb{R}^{k \times k}\) and a diagonal matrix \(D \in \mathbb{R}^{k \times k}\) s.t. \[A^TLA = Q^TDQ\]
    Which can be written as \[(AQ^T)^TL(AQ^T) = D\]
    Denoting by \(d_1, \dots, d_k\) the diagonal values of \(D\), the last equation yields \[\sum_{i=1}^k d_i = R_L(AQ^T) = R_L(A) = \sum_{i=1}^k \lambda_i\]
    Note that based on Remark \ref{remark:rayleigh} \(\lambda_i \leq d_i\) for each \(i\), as \(AQ^T \in O(n, k)\). Hence, \(d_i = \lambda_i\), i.e., \[(AQ^T)^TL(AQ^T) = \Lambda\]
    As the eigendecomposition of a matrix is unique, this yields \(AQ^T = V\), which means \(A = VQ\).
\end{proof}

\section{Algorithm Layouts} \label{app:algo}
\begin{center}
\begin{minipage}{0.9\textwidth}
\begin{algorithm}[H]
    \caption{SpectralNet training \citep{spectralnet}}\label{alg:spectralnet}
    \DontPrintSemicolon
    \KwIn{\(\mathcal{X} \subseteq \mathbb{R}^d\), number of dimensions \(k\), batch size \(m\)}
    \KwOut{Trained \(F_{\theta}\) which approximates the first \(k+1\) eigenfunctions up to isometry}

    Randomly initialize the network weights \(\theta\)\;
    \While{\(\mathcal{L}(\theta)\) not converged}{
        \textbf{Orthogonalization step:}\\
        Sample a random minibatch \(X\) of size \(m\)\;
        Forward propagate \(X\) and compute inputs to orthogonalization layer \(\Tilde{Y}\)\;
        Compute the \(QR\) factorization \(QR = \Tilde{Y}\)\;
        Set the weights of the orthogonalization layer to be \(\sqrt{m}R^{-1}\)\;
        \textbf{Gradient step:}\\
        Sample a random minibatch \(x_1,\dots, x_m\)\;
        Compute the \(m\times m\) affinity matrix \(W\)\;
        Forward propagate \(x_1,\dots, x_m\) to get \(y_1,\dots, y_m\)\;
        Compute the loss \(\mathcal{L}(\theta)\)(Sec. \ref{sec:spectralnet})\;
        Use the gradient of \(\mathcal{L}(\theta)\) to tune all \(F_{\theta}\) weights, except those of the output layer;
    }
\end{algorithm}
\end{minipage}
\end{center}

\begin{center}
\begin{minipage}{0.9\textwidth}
\begin{algorithm}[H]
    \caption{Eigenvectors separation}\label{alg:GrEASE}
    \DontPrintSemicolon
    \KwIn{\(\mathcal{X} \subseteq \mathbb{R}^d\), batch size \(m\), Trained \(F_{\theta}\) which approximates the first \(k+1\) eigenfunctions up to isometry}
    % \KwData{\(\mathcal{X} \subseteq \mathbb{R}^d\), batch size \(m\), Trained \(F_{\theta}\) which approximates the first \(k+1\) eigenfunctions up to isometry;}
    \KwOut{\(F_{\theta}\) which approximates the leading eigenfunctions}
    T \(\gets \lfloor\frac{|\mathcal{X}|}{m}\rfloor\)\;
    sample T minibatches \(X_i\in\mathbb{R}^{m\times d}\)\;
    Forward propogate all \(X_i\) and obtain \(F_{\theta}\) outputs \(Y_i\in\mathbb{R}^{m\times k+1}\)\;
    Compute the \(m \times m\) affinity matrices \(W_i\)\;      
    compute all corresponding RW-Laplacians \(L_i\)\;
    \(\Tilde{\Lambda} \gets \frac{1}{\text{T}}\sum_{i}Y_i^TL_iY_i\)\;
    Diagonalize \(\Tilde{\Lambda}\) to get \(\Tilde{Q}^T\) and the leading eigenvalues\;
    Sort the leading eigenvalues, and the columns of \(\Tilde{Q}^T\) correspondingly\;
    \(Q^T \gets\) last \(k\) columns of \(\Tilde{Q}^T\)\;
    To obtain the representation of a new test sample \(x_i\), compute \(y_i = F_{\theta}(x_i)Q^T\)
\end{algorithm}
\end{minipage}
\end{center}

\section{Implementation's Additional Considerations}
\label{app:additional_considerations}

\subsection{Time and Space Complexity}
Specifying the exact complexity of the method is difficult, As this is a non-convex optimization problem,
However, we can discuss the following approximate complexity analysis. Assuming constant input and output dimensions and a given network architecture, we can take a general view on the complexity of each iteration by the batch size \(m\). The heaviest computational operations at each iteration are the nearest-neighbors search, the QR decomposition and the loss computation (i.e., computation of the Rayleigh Quotient). For the nearest-neighbor search, we can use approximation techniques (e.g, LSH \cite{gionis1999similarity}) which work in almost linear complexity by \(m\). A naive implementation of the QR decomposition would lead to an \(\mathcal{O}(m^2)\) time complexity. The loss computation also takes \(\mathcal{O}(m^2)\) due to the required matrix multiplication. Thereby, the complexity of each iteration is quadratic by the batch size. This is comparable to other approximation techniques such as LOBPCG \cite{benner2011locallyLOBPCG} (which also utilizes sparse matrix operations techniques for faster implementation). However, \ourmethod\ leverages stochastic training, allowing each iteration to consider only a batch of the data, rather than the entire dataset.

Assessing the complexity of each epoch is now straightforward, and results a time complexity of \(\mathcal{O}(nm)\), where \(n\), the number of samples, satisfies \(n \gg m\). This indicates an almost-linear complexity.

\subsection{Graph Construction}
\label{app:graph}
To best capture the structure of the input manifold \(\mathcal{D}\), given by a finite number of samples \(\mathcal{X}\), we use a similar graph construction method used by Gomez et al. in UMAP \citep{umap}, proven to capture the local topology of the manifold at each point. However, as opposed to the method in \citep{umap}, \ourmethod\ does not compute the graph of all points, which can lead to scalability hurdles and impose significant memory demands. Instead, \ourmethod\ either computes small graphs on each batch, or can be provided by the user with an affinity matrix \(W\) corresponding to \(\mathcal{X}\). Our practical construction of the graph affinity matrix \(W\) is as follows:

Given a distance measure \(\delta\) between points, we first compute the \(k\)-nearest neighbors of each point \(x_i\) under \(\delta\), \(\{x_{i_1}, \dots, x_{i_k}\}\), and denote \[\rho_i = \min_j\delta(x_i, x_{i_j}),\ \sigma_i = \text{median}\{\delta(x_i, x_{i_j}) | 1 \leq j \leq k\}\]

Second, we compute the affinity matrix using the Laplace kernel 
\begin{equation*}
    W_{ij} = \begin{cases}
    \exp{\big( \frac{\rho_i-\delta(x_i, x_j)}{\sigma_i} \big)} & x_j \in \{x_{i_1}, \dots, x_{i_k}\} \\
    0 & \text{otherwise}
\end{cases}
\end{equation*}
Third, we symmetries \(W\) simply by taking \(\frac{W + W^T}{2}\).

We refer the reader to \cite{umap} for further discussion about the graph construction.

% \section{Evaluation Metrics}
\section{Grassmann Score}
\label{app:grassmann_score}
In this section, we provide the formulation for the Grassmann Score (GS) evaluation method, and present simple examples to visualize its meaning.

\subsection{Formalization of GS}
Grassmann distance (see Def. \ref{def:grassmann_distance}) is a metric function between equidimensional linear subspaces, where each is represented by an orthogonal matrix containing the basis as its columns. In other words, this is a metric which is invariant under multiplication by an orthogonal matrix.

\begin{definition} \label{def:grassmann_distance}
    Given two orthogonal matrices \(A, B \in \mathbb{R}^{n\times k}\), the Grassmann Distance between them is defined as: \[d_{Gr}(A, B) = \sum_{i=1}^k \text{sin}^2\theta_i\]
    where \(\theta_i = \arccos\sigma_i(A^T B)\) is the \(i\)th principal angle between \(A\) and \(B\), and \(\sigma_i\) is the \(i\)th smallest singular value of \(A^TB\).
\end{definition}

Assuming we are given a dataset \(\mathcal{X} = \{x_1, \dots, x_n\} \subseteq \mathbb{R}^d\) and a corresponding low-dimensional representation \(\mathcal{Y} = \{y_1, \dots, y_n\} \subseteq \mathbb{R}^k\). We want to evaluate the dissimilarity between the \textit{global structures} of \(\mathcal{X}\) and \(\mathcal{Y}\). We build graphs from \(\mathcal{X}\) and \(\mathcal{Y}\), saved as affinity matrices \(W_{\mathcal{X}}\) and \(W_{\mathcal{Y}}\), respectively. We construct the corresponding Unnormalized Laplacians (see Sec. \ref{sec:pre_se}) \(L_\mathcal{X}\) and \(L_\mathcal{Y}\). We define the matrices \(V_{\mathcal{X}}, V_{\mathcal{Y}} \in \mathbb{R}^{n\times t}\) so that their columns are the first \(t\) eigenvectors of \(L_\mathcal{X}, L_\mathcal{Y}\), respectively.

Finally, we define the GS of \(\mathcal{Y}\) (w.r.t \(\mathcal{X}\)) as follows:
\begin{definition} \label{def:grassmann_score}
    \(GS_{\mathcal{X}}(\mathcal{Y}) = d_{Gr}(V_{\mathcal{X}}, V_{\mathcal{Y}})\)
\end{definition}

\(t\) is a hyper-parameter of GS. A reasonable choice would be to take \(t=2\), which is equivalent to measure the Grassmann distance between the Fiedler vectors of the Laplacians. The Fiedler vector is known for its hold of the most important global properties. The larger \(t\), the more complicated structures are taken into consideration in the GS computation (which is not necceray desired).

Note that for the construction of the affinity matrices \(W_{\mathcal{X}}, W_{\mathcal{Y}}\) we use the same construction scheme detailed in App. \ref{app:graph}. This construction method is similar to the one presented by \cite{umap}, and proved to capture the local topology of the underlying manifold.

It is important to note that GS might ignore the local structures, while concentrating on the global structures (especially for smaller values of \(t\)). The ultimate goal in visualization is to find a balance between the global and local structure.

\subsection{Additional GS examples}
\begin{figure}[ht]
    \centering
    
    % First subfigure
    \begin{subfigure}[b]{0.9\textwidth}
        \centering
        \includegraphics[width=\textwidth]{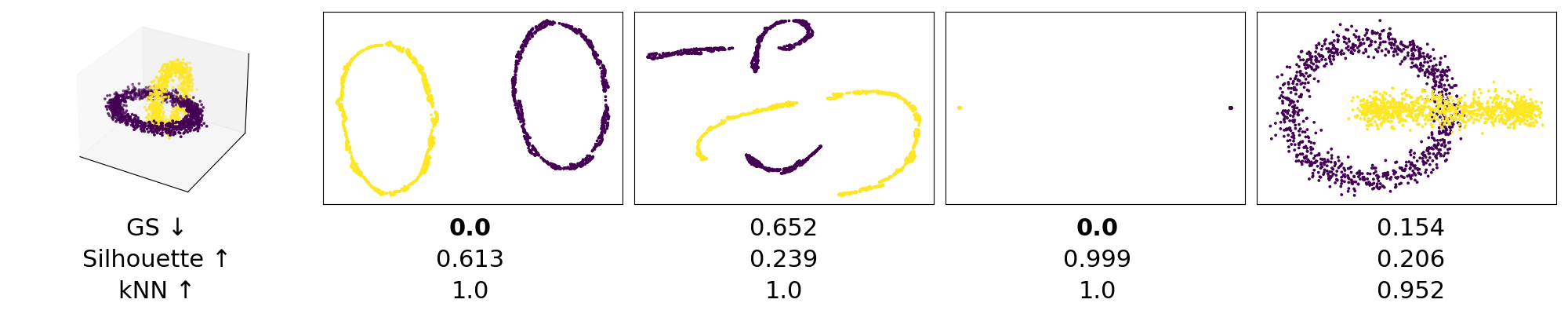}
        \caption{}
        \label{fig:gs_2circles}
    \end{subfigure}
    
    \vspace{0.5cm} % Space between subfigures
    
    % Second subfigure
    \begin{subfigure}[b]{0.9\textwidth}
        \centering
        \includegraphics[width=\textwidth]{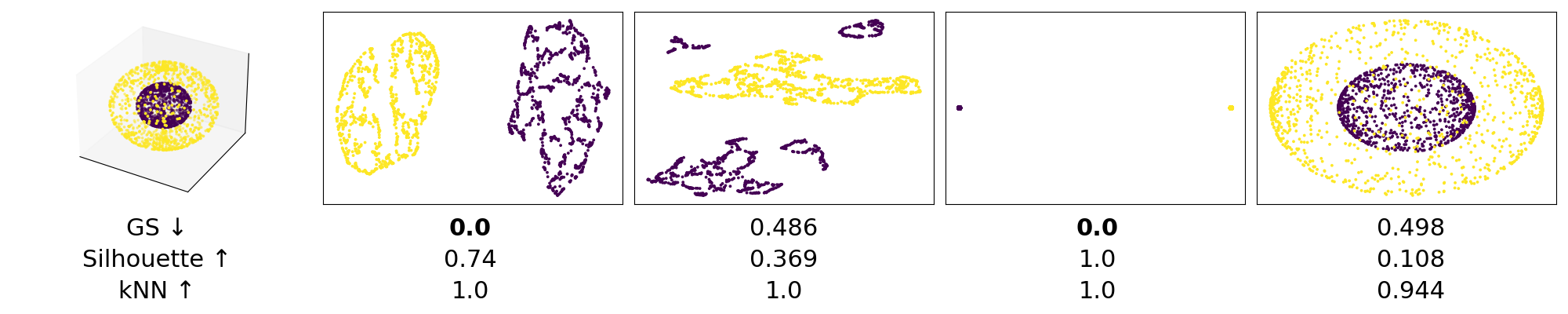}
        \caption{}
        \label{fig:gs_sphereInSphere}
    \end{subfigure}
    
    \caption{Additional demonstrations of the alignment between the intuitive expectation and the GS results on two toy dataset. Four possible 2-dimensional embeddings of these dataset are provided, along with their corresponding GS, kNN accuracy and Silhouette score. Unlike kNN and Silhouette, GS effectively captures the preservation of global structure.}
    \label{fig:gs_appendix}
\end{figure}

Fig. \ref{fig:gs_appendix} depicts two additional demonstrations of the alignment between the intuitive expectation and the GS results on two toy dataset. The basic global structure of both of these datasets is two distinct clusters. This structure is indeed captured by GS. However, kNN gives perfect score also when the one of the clusters is separated. Silhouette score favourites the 2-points embedding. Namely, it trade-offs local structure (i.e., giving lower score for preserving local structure, even when the global properties are the same).

% \section{Fine-Tuning \ourmethod\ with UMAP loss}
\section{NUMAP Ablation Study}
\label{app:numap_ft}
In this section, we start by considering two ablations of NUMAP, which for convenience we name NUMAP-SN and NUMAP-FT. NUMAP-SN is a replication of NUMAP's pipeline with SpectralNet replacing Sep-SpectralNet. That is, we show that eigenvector separation is necessary for NUMAP.

As for NUMAP-FT, this name refers to another method to get a generalizable version of UMAP by an extension of \ourmethod\, by fine-tuning the network with UMAP loss. We tried that idea, but were forced to stop this direction, as we stumbled upon the well-known catastrophic forgetting case.

Tab. \ref{tab:NUMAP_local_results_ablation} extends Tab. \ref{tab:NUMAP_local_results} with the results of NUMAP-SN and NUMAP-FT. Notably, NUMAP-SN local structure preservation are comparable with NUMAP and P. UMAP, as expected, while NUMAP-FT consistently fails in local structure preservation.

Tab. \ref{tab:NUMAP_global_results_ablation} extends Tab. \ref{tab:NUMAP_global_results} with the results of NUMAP-SN and NUMAP-FT. Notably, NUMAP-SN is less consistent than NUMAP in global structure preservation, resulting in worse performance on the Cifar10 and Banknote datasets. NUMAP-FT fails in global structure preservation on the Cifar10 and Appliances datasets. Although it achieves better GS on the Wine dataset, and comparable global structure preservation on the Banknote dataset, it compromises a lot on local structure preservation (see Tab. \ref{tab:NUMAP_local_results_ablation}).

\begin{figure}[htbp]
    \centering
    \includegraphics[width=0.82\textwidth]{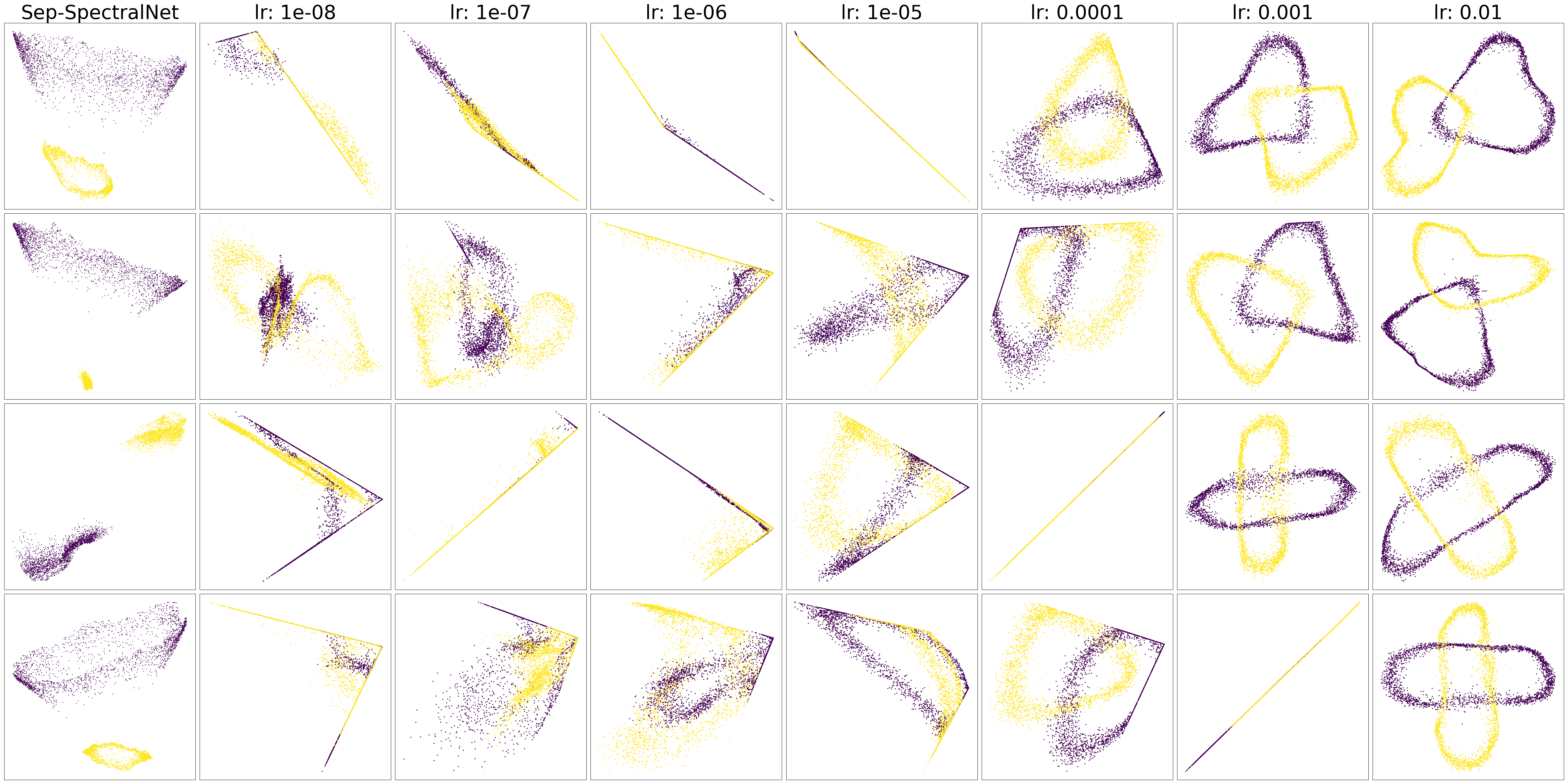}
    \caption{The catastrophic forgetting phenomenon when fine-tuning \ourmethod\ to match UMAP performance on the 2circles dataset. Each column represents a fine-tuning using a different learning-rate. Each row is a repetition, initialized with a different seed.}
    \label{fig:numap_ft}
\end{figure}

Figure \ref{fig:numap_ft} presents an experiment on the simple 2circles dataset. Each row is represented the same experiment, run with a different seed. We trained \ourmethod\ to output the 2D SE of the 2circles dataset, as shown in the left column. Then, we initialized a new network, with the same architecture, with the pre-trained weights from \ourmethod. This network was trained with UMAP loss, as in \citep{sainburg2021parametric}.  We tried different learning-rates for fine-tuning, to best match the desired UMAP embedding (i.e. retaining the local structure), without losing the global structure (e.g., separation of the two clusters). Unfortunately, there was no learning-rate that matched our goals.

\begin{table}[tbhp]
\scriptsize
\begin{center}
\begin{tabular}{llllll}
\multirow{1}{*}{{Metric}} & \multirow{1}{*}{\bf {Method}} & {Cifar10} & {Appliances} & {Wine} & {Banknote} \\
\hline\\
\multirow{4}{*}{{kNN \(\uparrow\)}}

& {P. UMAP} & {0.908\(_{\pm0.004}\)} & {-} & {0.953\(_{\pm0.033}\)} & {0.927\(_{\pm0.023}\)} \\

& {NUMAP (ours)} & {0.905\(_{\pm0.004}\)} & {-} & {0.950\(_{\pm0.024}\)} & {0.986\(_{\pm0.004}\)} \\

& {NUMAP-SN} & {0.903\(_{\pm0.002}\)} & {-} & {0.956\(_{\pm0.028}\)} & {0.963\(_{\pm0.032}\)} \\

& \bf {NUMAP-FT} & {0.577\(_{\pm0.081}\)} & {-} & {0.364\(_{\pm0.053}\)} & {0.686\(_{\pm0.046}\)} \\

\hline
\end{tabular}
\end{center}
\caption{\textbf{Local structure preservation ablation.} An extension of Tab. \ref{tab:NUMAP_local_results} with NUMAP-SN and NUMAP-FT.}
\label{tab:NUMAP_local_results_ablation}
\end{table}

\begin{table}[tbhp]
\scriptsize
\begin{center}
\begin{tabular}{llllll}
\multirow{1}{*}{\bf {Metric}} & \multirow{1}{*}{\bf {Method}} & {Cifar10} & {Appliances} & {Wine} & {Banknote} \\
\hline\\
\multirow{4}{*}{{GS \(\downarrow\)}}
& {P. UMAP} & {0.133\(_{\pm0.069}\)} & {0.710\(_{\pm0.293}\)} & {0.549\(_{\pm0.180}\)} & {0.722\(_{\pm0.079}\)} \\

& {NUMAP (ours)} & {0.054\(_{\pm0.021}\)} & {0.261\(_{\pm0.020}\)} & {0.429\(_{\pm0.124}\)} & {0.618\(_{\pm0.131}\)} \\

& {NUMAP-SN} & {0.281\(_{\pm0.357}\)} & {0.262\(_{\pm0.038}\)} & {0.441\(_{\pm0.182}\)} & {0.780\(_{\pm0.197}\)} \\

& {NUMAP-FT} & {0.680\(_{\pm0.334}\)} & {0.348\(_{\pm0.292}\)} & {0.002\(_{\pm0.001}\)} & {0.635\(_{\pm0.141}\)} \\

\hline
\end{tabular}
\end{center}
\caption{\textbf{Global structure preservation ablation.} An extension of Tab. \ref{tab:NUMAP_global_results} with NUMAP-SN and NUMAP-FT.}
\label{tab:NUMAP_global_results_ablation}
\end{table}

Another important question is the necessity of the residual connections. For that, we conduct another ablation study on the Cifar10 dataset. The results are shown in Tab. \ref{tab:NUMAP_rescon_local} and Tab. \ref{tab:NUMAP_rescon_global}. Notably, while the skip connections does not improve local structure preservation, they are crucial for global structure preservation, indicated by the lower GS.

\begin{table}[tbhp]
\scriptsize
\begin{center}
\color{blue}
\begin{tabular}{lll}
\multirow{1}{*}{{Metric}} & \multirow{1}{*}{\bf {Method}} & {Cifar10} \\
\hline\\
\multirow{2}{*}{{kNN \(\uparrow\)}}

& {NUMAP (no residual connections)} & {0.907\(_{\pm0.003}\)} \\

& {NUMAP} & {0.905\(_{\pm0.004}\)} \\

\hline
\end{tabular}
\end{center}
\caption{Ablation on residual connections for local structure preservation shows comparable performance with and without their use.}
\label{tab:NUMAP_rescon_local}
\end{table}

\begin{table}[tbhp]
\scriptsize
\begin{center}
\color{blue}
\begin{tabular}{lll}
\multirow{1}{*}{\bf {Metric}} & \multirow{1}{*}{\bf {Method}} & {Cifar10} \\
\hline\\
\multirow{2}{*}{{GS \(\downarrow\)}}
& {NUMAP (no residual connections)} & {0.191\(_{\pm0.293}\)} \\

& {NUMAP} & \bf {0.054\(_{\pm0.021}\)} \\

\hline
\end{tabular}
\end{center}
\caption{Ablation on residual connections for global structure preservation shows improved performance when they are used.}
\label{tab:NUMAP_rescon_global}
\end{table}

\section{Additional results}
\begin{figure}[t]
    \centering
    \includegraphics[width=0.76\linewidth]{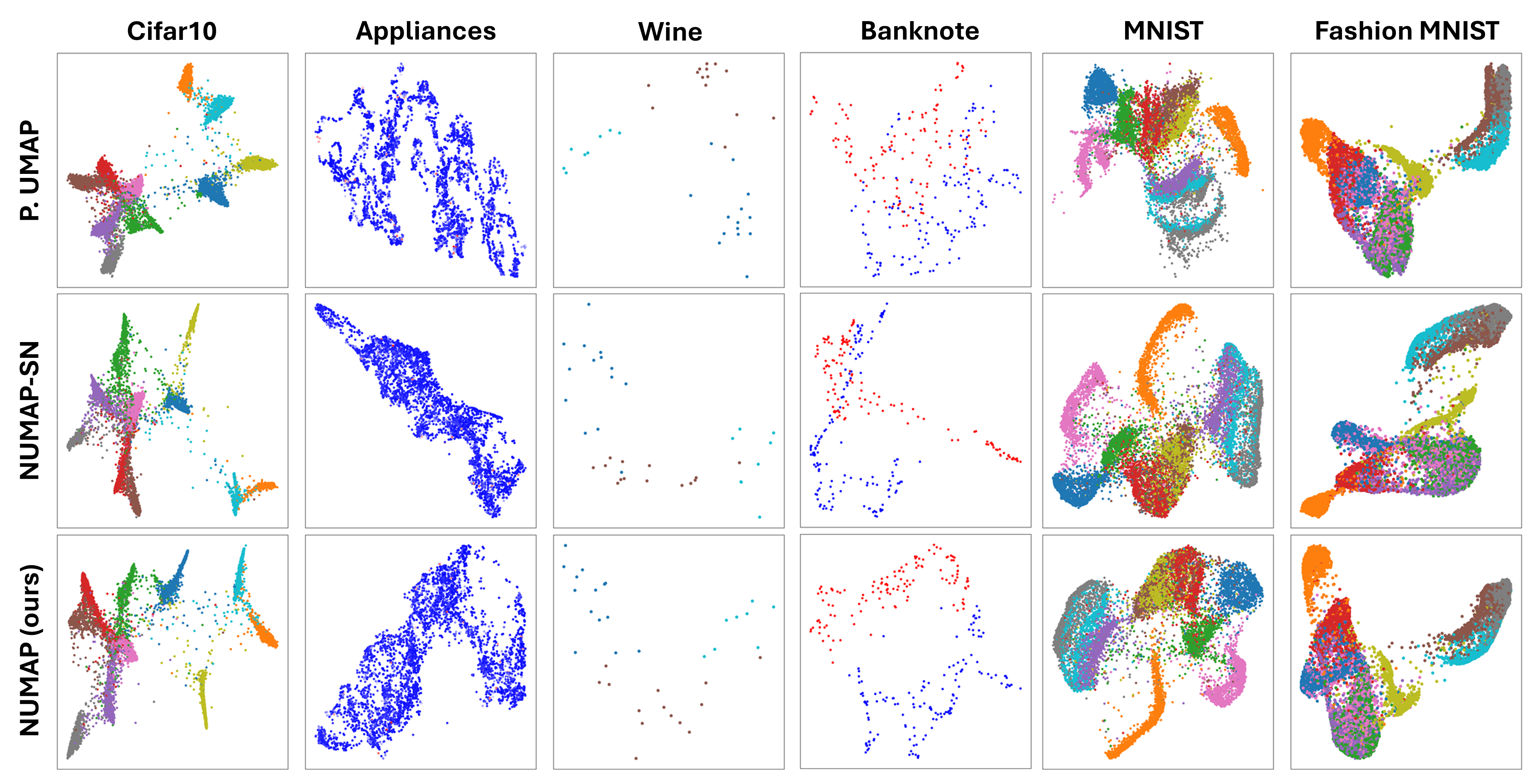}
    \caption{{Test set Visualizations of NUMAP, NUMAP-SN and P. UMAP on the datasets corresponding to Tab. \ref{tab:NUMAP_local_results} and Tab. \ref{tab:NUMAP_global_results}.}}
    \label{fig:visus}
\end{figure}

\begin{figure}[htbp]
    \centering
    \includegraphics[width=\linewidth]{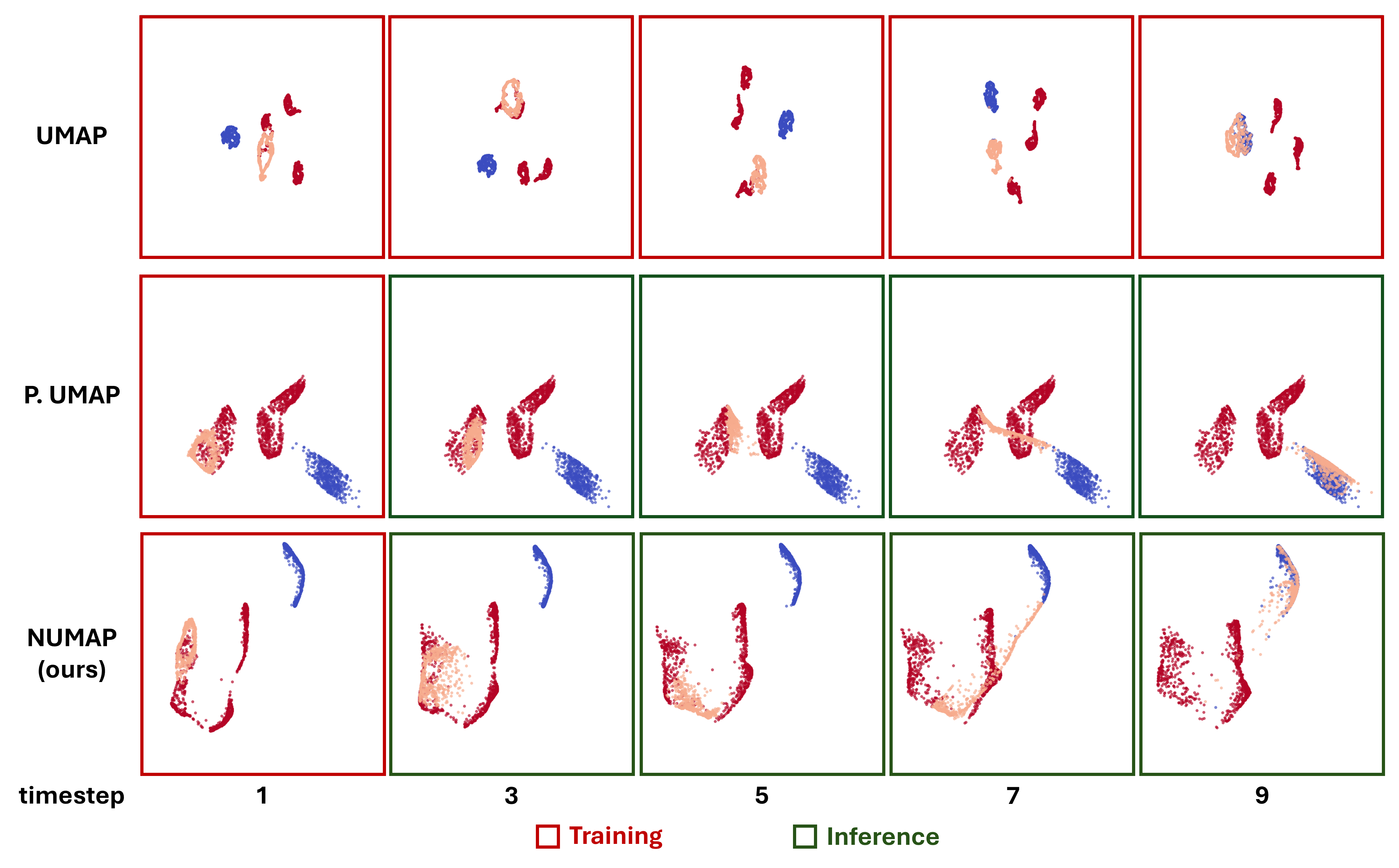}
    \caption{{Fig. \ref{fig:numap_teimesteps} extension with P. UMAP.}}
    \label{fig:teimesteps_extended}
\end{figure}

The full results of Fig. \ref{fig:GrEASE_results} are summarized in Tab. \ref{tab:GrEASE_results}. The visualizations corresponding to Tab. \ref{tab:NUMAP_global_results} are depicted in Fig. \ref{fig:visus}. Fig. \ref{fig:teimesteps_extended} extends Fig. \ref{fig:numap_teimesteps} with P. UMAP.

Tab. \ref{tab:NUMAP_local_results_extended} extends Tab. \ref{tab:NUMAP_local_results_ablation} with additional two datasets: MNIST \citep{mnist} and FashionMNIST \citep{xiao2017fashionMNIST}. Tab. \ref{tab:NUMAP_global_results_extended} correspondingly extends Tab. \ref{tab:NUMAP_global_results_ablation}.

\begin{table}[t]
\scriptsize
\caption{A comparison between \ourmethod\ and SpectralNet dimensional SE and Fiedler Vector (FV) approximation on real-world datasets. The values are the mean and standard deviation of the \(\text{sin}^2\) distance between the predicted and true eigenvector, over 10 runs. Lower is better. \ourmethod\ ability to separate the eigenvectors is evident.}
\label{tab:GrEASE_results}
\begin{center}

% First big table
\begin{tabular}{llllll}
\multirow{1}{*}{Dataset}  & \multirow{1}{*}{Method} & \multicolumn{1}{c}{\bf \(v_2\)}  & \multicolumn{1}{c}{\bf \(v_3\)}    & \multicolumn{1}{c}{\bf \(v_4\)}    & \multicolumn{1}{c}{\bf \(v_5\)} \\
\hline\\
\multirow{2}{*}{Cifar10} & \ourmethod & \(0.016_{\pm0.004}\) & \(0.052_{\pm0.008}\) & \(0.069_{\pm0.034}\) & \(0.106_{\pm0.037}\) \\
 & SpectralNet & \(0.449_{\pm0.199}\) & \(0.325_{\pm0.148}\) & \(0.399_{\pm0.194}\) & \(0.414_{\pm0.17}\) \\
\hline
\multirow{2}{*}{Appliances} & \ourmethod & \(0.063_{\pm0.002}\) & \(0.094_{\pm0.007}\) & \(0.109_{\pm0.001}\) & - \\
 % & SpectralNet & \(0.381_{\pm0.093}\) & \(0.429_{\pm0.122}\) & \(0.358_{\pm0.155}\) & - \\
& SpectralNet & \(0.307_{\pm0.047}\) & \(0.530_{\pm0.114}\) & \(0.401_{\pm0.106}\) & - \\
\hline
\multirow{2}{*}{KMNIST} & \ourmethod & \(0.0.044_{\pm0.002}\) & \(0.101_{\pm0.010}\) & - & - \\
& SpectralNet & \(0.372_{\pm0.174}\) & \(0.396_{\pm0.137}\) & - & - \\
\hline
\multirow{2}{*}{Parkinsons} & \ourmethod & \(0.056_{\pm0.006}\) & - & - & - \\
& SpectralNet & \(0.229_{\pm0.138}\) & - & - & - \\
\hline
\end{tabular}

\end{center}
\end{table}

\begin{table}[tbhp]
\scriptsize
\begin{center}
\begin{tabular}{llllllll}
\multirow{1}{*}{\bf {Metric}} & \multirow{1}{*}{\bf {Method}} & {Cifar10} & {Appliances} & {Wine} & {Banknote} & {Mnist} & {FashionMnist} \\
\hline\\
\multirow{4}{*}{{kNN \(\uparrow\)}}

& {P. UMAP} & {0.908\(_{\pm0.004}\)} & {-} & {0.953\(_{\pm0.033}\)} & {0.927\(_{\pm0.023}\)} & {0.801\(_{\pm0.010}\)} & {0.717\(_{\pm0.006}\)} \\

& {NUMAP (ours)} & {0.905\(_{\pm0.004}\)} & {-} & {0.950\(_{\pm0.024}\)} & {0.986\(_{\pm0.004}\)} & {0.758\(_{\pm0.009}\)} & {0.695\(_{\pm0.003}\)} \\

& {NUMAP-SN} & {0.903\(_{\pm0.002}\)} & {-} & {0.956\(_{\pm0.028}\)} & {0.963\(_{\pm0.032}\)} & {0.750\(_{\pm0.010}\)} & {0.695\(_{\pm0.006}\)} \\

& {NUMAP-FT} & {0.577\(_{\pm0.081}\)} & {-} & {0.364\(_{\pm0.053}\)} & {0.686\(_{\pm0.046}\)} & {0.329\(_{\pm0.103}\)} & {0.153\(_{\pm0.035}\)} \\

\hline
\end{tabular}
\end{center}
\caption{An extension of Tab. \ref{tab:NUMAP_local_results_ablation} with the MNIST and FashionMNIST datasets. Local preservation results are corresponding to Tab. \ref{tab:NUMAP_local_results_ablation}. Namely, NUMAP, P. UMAP and NUMAP-SN are comparable, with NUMAP-FT failing to preserve local structure.}
\label{tab:NUMAP_local_results_extended}
\end{table}

\begin{table}[tbhp]
\scriptsize
\begin{center}
\begin{tabular}{llllllll}
\multirow{1}{*}{\bf {Metric}} & \multirow{1}{*}{\bf {Method}} & {Cifar10} & {Appliances} & {Wine} & {Banknote} & {Mnist} & {FashionMnist} \\

\hline\\
\multirow{4}{*}{{GS \(\downarrow\)}}

& {P. UMAP} & {0.133\(_{\pm0.069}\)} & {0.710\(_{\pm0.293}\)} & {0.549\(_{\pm0.180}\)} & {0.722\(_{\pm0.079}\)} & {0.311\(_{\pm0.059}\)} & {0.029\(_{\pm0.006}\)} \\

& {NUMAP (ours)} & {0.054\(_{\pm0.021}\)} & {0.261\(_{\pm0.020}\)} & {0.429\(_{\pm0.124}\)} & {0.618\(_{\pm0.131}\)} & {0.304\(_{\pm0.042}\)} & {0.032\(_{\pm0.022}\)} \\

& {NUMAP-SN} & {0.281\(_{\pm0.357}\)} & {0.262\(_{\pm0.038}\)} & {0.441\(_{\pm0.182}\)} & {0.780\(_{\pm0.197}\)} & {0.405\(_{\pm0.199}\)} & {0.036\(_{\pm0.016}\)} \\

& {NUMAP-FT} & {0.680\(_{\pm0.334}\)} & {0.348\(_{\pm0.292}\)} & {0.002\(_{\pm0.001}\)} & {0.635\(_{\pm0.141}\)} & {0.411\(_{\pm0.153}\)} & {0.272\(_{\pm0.146}\)} \\
\hline
\end{tabular}
\end{center}
\caption{An extension of Tab. \ref{tab:NUMAP_global_results_ablation} with the MNIST and FashionMNIST datasets. Global preservation results on MNIST are comparable between UMAP and P. UMAP, while superior to NUMAP-SN and NUMAP-FT. Regarding FashionMNIST, NUMAP, FashionMNIST, and NUMAP-SN are comparable, with NUMAP-FT having worse preservation.}
\label{tab:NUMAP_global_results_extended}
\end{table}

\section{Technical Details}
\label{app:technical_details}

\begin{table}[]
    \centering
    \scriptsize
    \captionsetup{width=0.9\textwidth}
    \caption{Technical details of the real-world datasets used for \ourmethod\ and NUMAP experiments.}
    \begin{tabular}{ccccccccc}
         & Cifar10  & Appliances & KMNIST &  Parkinsons & Wine & Banknote & MNIST & FashionMNIST\\
         \hline
         \#samples & 60,000 & 19735 & 70,000 & 5875 & 178 & 1372 & 60,000 & 60,000 \\
         \#features & 500 & 28 & 784 & 19 & 13 & 4 & 784 & 784 \\
    \end{tabular}
    \label{tab:tech}
\end{table}

\begin{table}[]
    \centering
    \captionsetup{width=0.9\textwidth}
    \caption{Technical details in the \ourmethod\ experiments for all datasets.}
    \begin{tabular}{cccccc}
         & Moon & Cifar10  & Appliances & KMNIST &  Parkinsons\\
         \hline
         Batch size & 2048 & 2048 & 2048 & 2048 & 512 \\
         n\_neighbors & 20 & 20 & 20 & 20 & 5 \\
         Initial LR & \(10^{-2}\) & \(10^{-2}\) & \(10^{-3}\) & \(10^{-3}\) & \(10^{-2}\)\\
         Optimizer & ADAM & ADAM & ADAM & ADAM & ADAM\\
    \end{tabular}
    \label{tab:tech}
\end{table}

To compute the ground truth SE on the train set and its corresponding eigenvalues, we constructed an affinity matrix \(W\) from the train set (as detailed in Appendix \ref{app:graph}), with a number of neighbors detailed in Table \ref{tab:tech}. After constructing \(W\), we computed the leading \(k\) eigenvectors of its corresponding Unnormalized Laplacian \(L = D - W\) via Python's Numpy SVD or SciPy LOBPCG SVD (depending on the size). To get the generalization ground truth, we constructed an affinity matrix \(W\) from the train and test sets combined, computed the leading \(k\) eigenvectors of its corresponding Unnormalized Laplacian \(L = D - W\), and extracted the representations corresponding to the test samples. We used a train-test split of 80-20 for all datasets.

For the SE implementation via sparse matrix decomposition techniques, we used Python's sklearn.manifold.SpectralEmbedding, using a default configuration (in particular, 10 jobs, 1\% neighbors).

The architectures of \ourmethod's and SpectralNet's networks in all of the experiments were as follows: size = 256; ReLU, size = 256; ReLU, size = 512; ReLU, size = \(k+1\); orthonorm. NUMAP's second NN and PUMAP's NN architectures for all datasets was: size = 200; ReLU, size = 200; ReLU, size = 200; ReLU, size = 2; The SE dimensions for NUMAP were: Cifar10 - 10; Appliances - 5; Wine - 10; Banknote - 3; Mnist - 10, FashionMnist - 10. {For the datasets in Fig. \ref{fig:qualitative}, from top to bottom: Circles - 5, Cylinders - 11, Line - 2.}

The learning rate policy for \ourmethod\ and SpectralNet is determined by monitoring the loss on a validation set (a random subset of the training set); once the validation loss did not improve for a specified number of epochs, we divided the learning rate by 10. Training stopped once the learning rate reached \(10^{-7}\). In particular, we used the following approximation to determine the patience epochs, where \(n\) is the number of samples and \(m\) is the batch size: if \(\frac{n}{m} \leq 25\), we chose the patience to be 10; otherwise, the patience decreases as \(\max{(1, \frac{250m}{n})}\) (i.e., the number of iterations is the deciding feature).

To run UMAP, we used Python's umap-learn implementation (UMAP's formal implementation). We used the built-in initialization option "spectral" (i.e., SE), and initialized contumely with PCA (implemented via Python's sklearn.decomposition.PCA) and \ourmethod. For Parametric UMAP we used the Pytorch implementaion \citep{umap_pytorch}. For all methods we used a default choice of 10 neighbors.

As for the evaluation methods, we used a default choice of 5 neighbors to compute the kNN accuracy. The graph construction for GS is as detailed in App. \ref{app:graph}, using 50 neighbors to ensure connectivity.

For the computation of the p-values in Tab. \ref{tab:NUMAP_global_results} we used an independent t-test to compare the means of the 10 GS results, that were obtained from 10 runs using 10 different seeds.

\paragraph{Time-series simulation.} We simulated two complex distributions in a 10-dimensional space. At each of the ten time steps, we sample a total of 5000 data points, 25\% of which belong to the dynamic distribution (visualized by the pink dots in Fig. \ref{fig:numap_teimesteps}), while the other two distributions are kept the same. The dynamic distribution starts at the first (red) distribution, and linearly transitions into the other (blue). We used UMAP default parameters settings to visualize each time-step separately. As for NUMAP, we trained only on the first two time-steps, and obtained the others using a simple feed-forward operation.

We ran the experiments using GPU: NVIDIA A100 80GB PCIe; CPU:  Intel(R) Xeon(R) Gold 6338 CPU @ 2.00GHz;

\end{document}